\documentclass{article}

\usepackage[preprint]{neurips_2026}

\usepackage[utf8]{inputenc}
\usepackage[T1]{fontenc}
\usepackage{hyperref}
\usepackage{url}
\usepackage{booktabs}
\usepackage{amsfonts}
\usepackage{nicefrac}
\usepackage{microtype}
\usepackage{xcolor}
\usepackage{graphicx}
\usepackage{array}
\usepackage{enumitem}
\usepackage{amsmath}
\usepackage{subcaption}
\usepackage{tcolorbox}

\usepackage{makecell}
\usepackage{multirow}

\newcommand{\red}[1]{{\color{red}#1}}
\newcommand{\blue}[1]{{\color{blue}#1}}
\newcommand{\jynew}[1]{{#1}}

\title{UniFunc3D: Unified Active Spatial-Temporal Grounding for 3D Affordance Segmentation}

\author{%
    Jiaying Lin \qquad \qquad Dan Xu\thanks{Corresponding Author.}\\
     Department of Computer Science and Engineering\\
  The Hong Kong University of Science and Technology\\
  \texttt{csjylin@gmail.com} \qquad  \texttt{danxu@cse.ust.hk} \\
}

\begin{document}

\maketitle

\begin{abstract}
Affordance segmentation in 3D scenes requires an agent to ground implicit natural-language instructions into precise masks of fine-grained interactive elements. Existing training-free methods typically rely on fragmented pipelines, which introduce visual blindness during task parsing and limit accuracy through single-scale spatial and temporal processing.
We present UniFunc3D, a unified and training-free framework that treats the multimodal large language model as an active observer. 
By utilizing a unified MLLM backbone,
UniFunc3D performs joint semantic-temporal-spatial reasoning to ground task decomposition in direct visual evidence. Our approach introduces active spatial-temporal grounding with a coarse-to-fine strategy. This allows the model to select correct video frames adaptively and focus on high-detail interactive parts while preserving the global context necessary for disambiguation.
On SceneFun3D, our UniFunc3D achieves state-of-the-art performance, surpassing prior training-free methods by a large margin with a relative 59.9\% mIoU improvement, and even outperforming training-based methods without any task-specific training. 
Code is available on our project page: \url{https://jiaying.link/unifunc3d}.
\end{abstract}

\section{Introduction}

For an embodied agent to operate effectively in human environments, it must look beyond simple object labels and understand affordances, the latent functional properties that enable interaction. While standard open-vocabulary 3D segmentation focuses on identifying what an object is (\textit{e.g.}, ``a cabinet''), affordance segmentation requires determining how to interact with it based on intent. For instance, given the command ``turn on the ceiling light,'' an agent must infer that the target is a specific wall switch, even if the word ``switch'' is never mentioned. This task is inherently difficult because it requires a synergy of high-level world knowledge for language interpretation and fine-grained spatial perception for localizing components at a small scale.

\begin{figure}[t]
    \centering
    \includegraphics[width=0.95\linewidth]{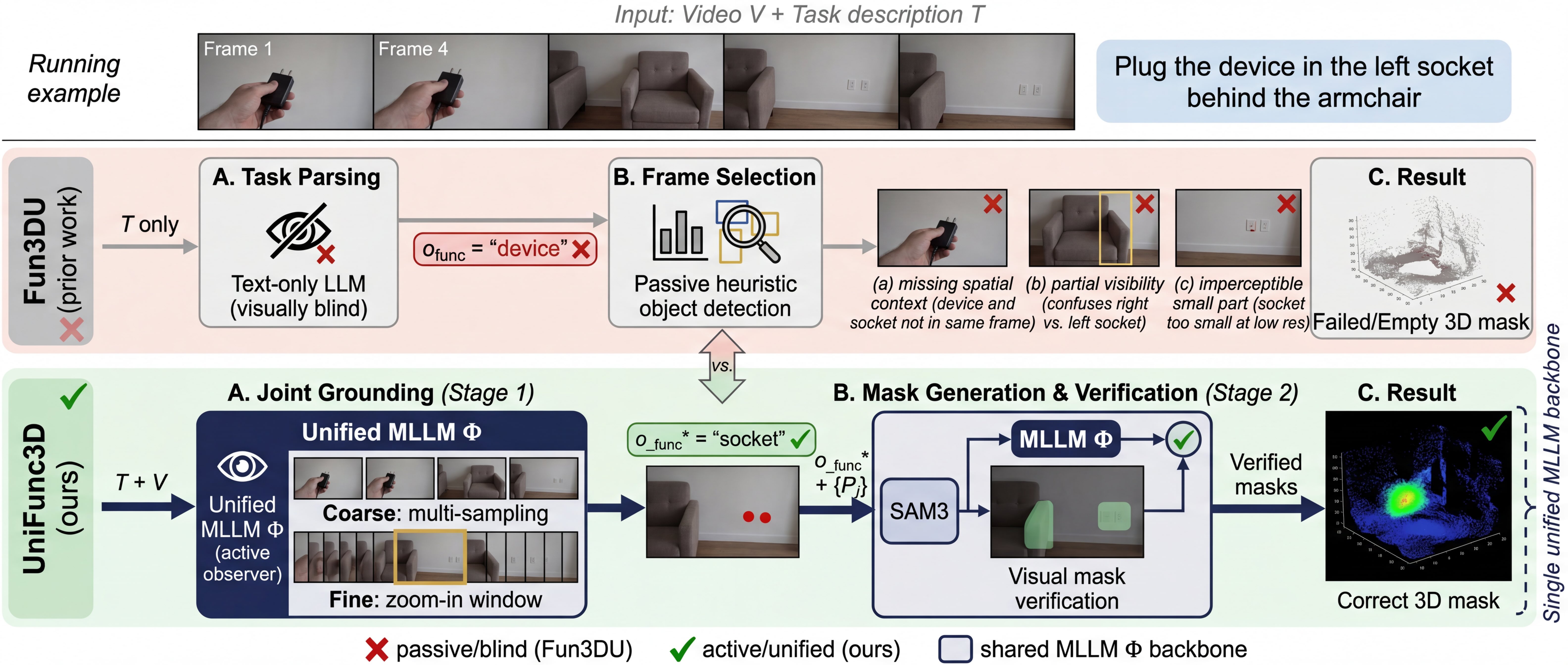}
    \caption{\textbf{Overview of UniFunc3D compared to existing fragmented pipelines.} (Top) Prior methods like Fun3DU rely on a visually blind text-only LLM for initial task parsing. Coupled with single-scale passive heuristic frame selection, this fragmented approach suffers from three critical failure modes: semantic misinterpretations (Task A$\rightarrow$ Task B), spatial-temporal context inconsistencies (a, b) and imperceptible small target (c), leading to error in the final output.
    (Bottom) Our proposed UniFunc3D addresses these limitations by utilizing a unified MLLM as an active observer. By employing a coarse-to-fine active spatial-temporal grounding strategy alongside visual mask verification, UniFunc3D consolidates semantic, temporal, and spatial reasoning in a unified MLLM. This allows the model to accurately ground implicit targets and generate precise fine-grained 3D functional masks while preserving necessary global context.}
    \label{fig:video-limitations}
\end{figure}

Existing training-free methods (\textit{e.g.}, Fun3DU~\cite{Functionality}) typically relies on fragmented pipelines that suffer from a fundamental lack of active spatial-temporal reasoning. These approaches often begin with a ``visually blind'' reasoning stage where a text-only LLM (\textit{i.e.}, LLaMA-3.1-9B~\cite{grattafiori2024llama}) to decompose the input text description into two kinds of objects: the contextual object (\textit{i.e.}, the object that contains or is related to the functional object) and the functional object (\textit{i.e.}, the ultimate object(s) or parts to segment).
Such a fragmented pipeline means that if one of these objects is wrongly detected, then the final result will be wrong.
However, as their first stage is text-only, their decomposition occurs without seeing the scene, leading to inaccurate contextual and functional identifications. For example, for the input \texttt{plug the device in the left socket behind the armchair}, Fun3DU may wrongly identify the \texttt{device} as the functional object while the correct one is \texttt{socket}.
Second, they utilize passive heuristic rules (\textit{e.g.}, hand-crafted, threshold-based scores that weight the centeredness and uniformity of detected contextual object masks) to select video frames for independent processing. They identify all frames containing the contextual object based purely on object category and assume the detected contextual object and the target functional part always reside in the same frame, an ideal situation that is frequently not guaranteed.
What is worse, these methods process images as isolated single frames and do not utilize temporal information from multiple frames, leading to errors when disambiguating spatial relationships or aggregating visibility across views.
Third, when processing images, they rely on single-scale processing and lack a human-like ``zoom-in'' mechanism. Because it cannot adaptively focus on important frames at a higher resolution, small functional parts occupy tiny regions that appear as imperceptible noise.


We observe that these issues are not just isolated errors but are symptoms of a ``perception-reasoning gap''. Because existing methods cannot actively look for the context they need, they fall victim to cascading errors where a mistake in the initial ``blind'' reasoning or heuristic frame selection irrecoverably ruins all downstream steps.
This motivates us to design a method that actively seeks out the necessary spatial-temporal context or aggregates multi-view context and multi-scale content to resolve fine-grained details for this challenging task.

To address these limitations, we introduce UniFunc3D, a unified and training-free framework that consolidates semantic, temporal, and spatial reasoning within a single multimodal large language model (MLLM).
In UniFunc3D, we propose an active spatial-temporal grounding process that eliminates handcrafted and heuristic rules with the heavy hyperparameter controls and unstable object detection found in Fun3DU. It observes the entire video to adaptively identify informative temporal segments and enables the selection of optimal candidate frames based on direct visual evidence.
During this process, the model jointly conducts semantic, temporal, and spatial reasoning to directly locate the target functional object among multiple input video frames. This unified architecture prevents cascading errors by allowing these distinct reasoning aspects to mutually inform and reinforce one another. This synergy is essential for resolving fine-grained interactive parts in complex 3D scenes.
Besides, our UniFunc3D utilizes a coarse-to-fine strategy that mimics human-like perception: In the coarse round, the model surveys the video at low resolution to identify candidates. In the fine round, it processes a dense temporal window at native high resolution, resolving spatial anchors and small parts while retaining complete scene context for disambiguation.
This coherent architecture ensures the agent can self-correct its initial estimates, avoiding the error propagation inherent in naive zoom-in pipelines with region-cropping.
Our key contributions include:
\begin{itemize}[leftmargin=*]
\item Unified Multimodal Architecture: We eliminate the cascading errors of fragmented pipelines by consolidating reasoning and perception into a single, spatial-temporal and visual-aware MLLM.
\item Active spatial-Temporal Grounding: We replace passive heuristics with a multi-sampling and verification strategy that allows the model to autonomously select the most informative content from video sequences.
\item Human-like Coarse-to-fine Perception: Our two-round approach achieves high precision on fine-grained elements without external cropping, preserving global context for robust spatial reasoning.
\item State-of-the-Art Performance: UniFunc3D achieves state-of-the-art results on SceneFun3D, largely surpassing both training-free methods and training-based methods, even though without task-specific training and large models.
\end{itemize}

\section{Related Work}
\label{sec:related}

\noindent \textbf{Affordance segmentation.}
Affordance understanding has evolved from object-centric methods~\cite{qian2024affordancellm,3d-affordancellm,grounding3d,lu2025geal} focusing on isolated objects, to scene-level reasoning in complex environments.
Early 2D affordance methods~\cite{xu2024weakly,li2024one,qian2024affordancellm} leverage segmentation models and weakly supervised learning for RGB-based affordance parsing but lack 3D grounding~\cite{wang2025n3d} necessary for embodied interaction.
Unlike previous 3D affordance datasets focusing on individual objects~\cite{deng20213daffnet} \jynew{or closed-vocabulary setting~\cite{articulate3d}}, SceneFun3D~\cite{delitzas2024scenefun3d} provides 230 high-resolution real-world indoor scenes with over 3,000 challenging task descriptions requiring world knowledge and spatial reasoning.
Fun3DU~\cite{Functionality} is the first dedicated method for this benchmark, employing a four-stage training-free pipeline: (1) a text-only LLM (\textit{i.e.}, LLaMA-3.1-9B) performs reasoning to identify contextual and functional objects; (2) open-vocabulary object segmentation locates contextual objects to select views; (3) a VLM grounds functional objects; and (4) geometric lifting aggregates 2D masks into 3D.
TASA~\cite{tasa} introduces task-aware frame selection and 3D geometric refinement with learnable components, requiring training on SceneFun3D to optimize for the task.
AffordBot~\cite{affordbot} operates directly on 3D point clouds rather than videos, rendering surround-view images and fine-tuning Mask3D~\cite{schult2023mask3d} for 3D instance segmentation. Both require a large MLLM (Qwen2.5-VL-72B).
However, these methods face critical limitations: Fun3DU operates initial reasoning without visual input, leading to errors in ambiguous cases. It processes frames independently without leveraging temporal context for spatial disambiguation.
Training-based methods like TASA and AffordBot require task-specific data and lack generalization to unseen domains. Specifically, they require point clouds as input and ground-truth point cloud annotations for training.

Instead, UniFunc3D uses a unified MLLM that jointly performs visual reasoning, temporal grounding, and spatial localization in a unified model, eliminating visual blindness and information loss across pipeline stages while remaining fully training-free.
\jynew{Note that there are additional related work sections in the Appendix due to the page limit.}

\section{Method}
\label{sec:method}

\subsection{Problem Formulation}
\label{sec:probform}

Given a 3D scene represented as a point cloud $\mathcal{C} = \{c_i\}_{i=1}^C$ where each $c_i \in \mathbb{R}^3$ is a 3D point, and a set of posed RGB-D views $\mathcal{V} = \{v_i\}_{i=1}^V$ captured from different viewpoints, along with a natural language task description $T$, our goal is to segment the functional object $F$ that enables task completion.
Unlike explicit queries that directly name objects (\textit{e.g.}, ``segment the handle''), 3D affordance segmentation requires inferring: (1) which object to interact with based on world knowledge, (2) which specific part enables the action, and (3) which instance among multiple similar objects satisfies spatial constraints in $T$.
We output a 3D mask $\mathcal{M}_F \subset \mathcal{C}$ indicating which points belong to the functional object.

\begin{figure}[t]
    \centering
    \includegraphics[width=\linewidth]{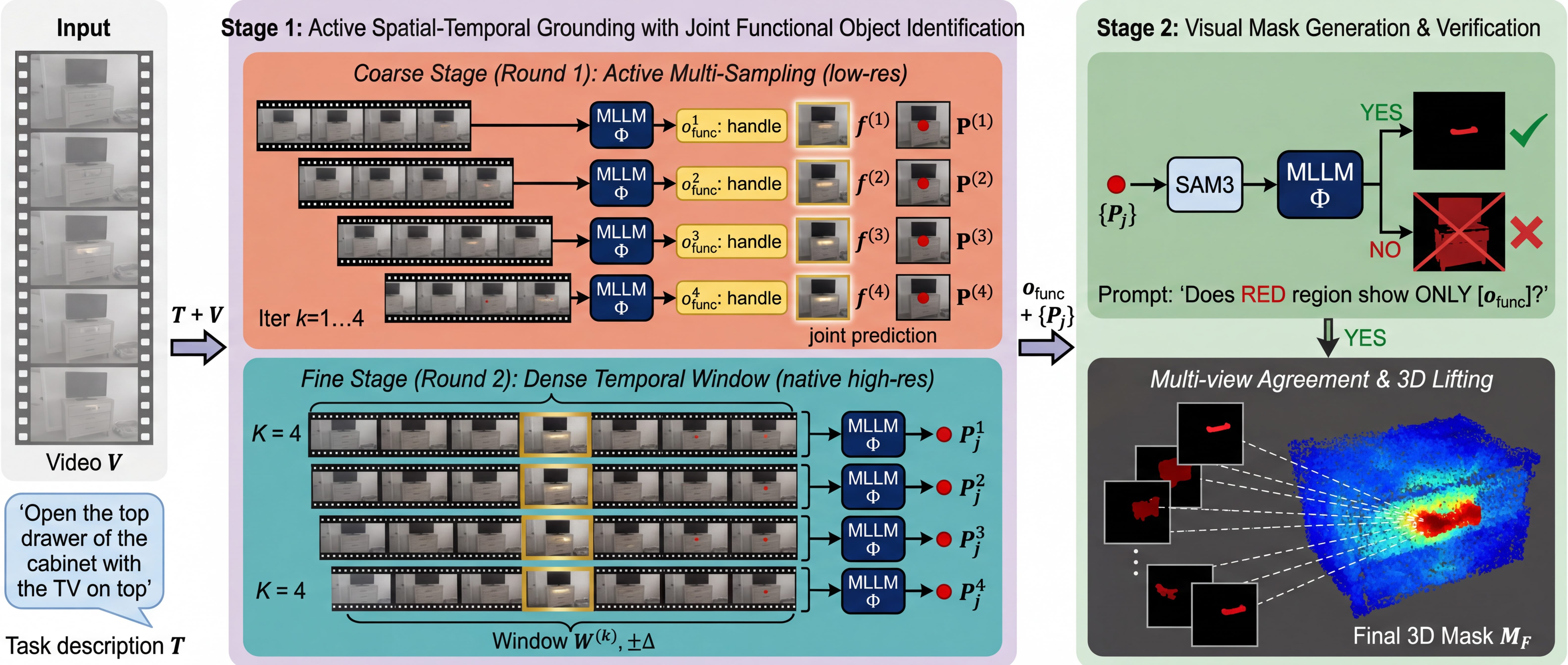}
\vspace{-6mm}
    \caption{\textbf{Method overview.} UniFunc3D employs a unified MLLM with \textit{active spatial-temporal grounding with joint functional object identification}: the coarse stage (Round 1) actively surveys low-resolution video frames across multiple sampling iterations and selects the most informative candidate via visual verification; the fine stage (Round 2) processes a dense temporal window at native high resolution, delivering zoom-in capability while preserving global scene context for precise localization.  \textit{Visual mask generation and verification} uses SAM3 for segmentation with MLLM-based mask verification, then multi-view 3D lifting to obtain the final 3D masks.}
    \label{fig:method}
\end{figure}



\subsection{Unified MLLM with Visually Grounded Reasoning}
\label{sec:unified_reasoning}

Prior work~\cite{Functionality} performs task decomposition using a text-only LLM, generating hypotheses about functional objects without visual verification.
This leads to errors when objects have ambiguous functional parts that cannot be inferred from language alone. Fig.~\ref{fig:method} illustrates the full pipeline.
We address this by applying the same MLLM $\Phi$ across two visually grounded stages:
\begin{equation*}
    (o_{\text{func}}^*,\, \{\mathbf{P}_j\}) = \Phi_{\text{ground}}(T,\, V),\qquad
    b_i^j = \Phi_{\text{verify}}(\tilde{I}_j,\, o_{\text{func}}^*)
\end{equation*}
where $\mathcal{V}$ are input video frames, $o_{\text{func}}^*$ is the functional object name jointly predicted by the grounding stage, $\mathbf{P}_j$ is the set of affordance points predicted in frame $I_j$, and $b_i^j \in \{\text{YES}, \text{NO}\}$ is the result of the verification for the mask $M_i^j$ rendered as an overlay image $\tilde{I}_j$.
The two stages form a reasoning chain in which each stage consumes visual evidence: \textbf{Stage 1 (Active spatial-temporal grounding)} jointly identifies \textit{what} to segment and \textit{where} it is located by simultaneously predicting the functional object name and grounding affordance points across video frames, as detailed in Sec.~\ref{sec:active_grounding}; \textbf{Stage 2 (Visual mask generation and verification)} generates target masks from the predicted points and visually inspects each segmentation overlay, filtering over-segmented predictions before 3D lifting, as detailed in Sec.~\ref{sec:segmentation}.



\subsection{Active Spatial-Temporal Grounding}
\label{sec:active_grounding}

Existing training-free methods like Fun3DU~\cite{Functionality} passively ranks frames by object detection scores for a pre-determined contextual object category, a strategy with no mechanism to adapt when spatial anchors are absent or when fine-grained parts demand higher-resolution inspection.

We address this with \textbf{active spatial-temporal grounding}: rather than relying on passive object-detection heuristics, the MLLM acts as an active observer that surveys multiple temporal slices of the video and selects the most informative frames through direct visual evidence, autonomously replacing all hand-crafted rules and hyperparameters.
To resolve the small-part visibility challenge without sacrificing global context, we adopt a \textbf{human-like coarse-to-fine perception} strategy. A naive crop-and-reprocess zoom-in agent introduces cascaded errors,
and an incorrect crop irrecoverably commits the pipeline to a wrong hypothesis. Instead, our coarse stage surveys the full video at low resolution to identify a candidate frame; the fine stage then processes a dense temporal window around it at native high resolution, delivering zoom-in capability while retaining the complete scene so the model can self-correct coarse estimates.

\noindent\textbf{Coarse stage (Round 1): Active multi-sampling frame selection.}
Given a video sequence $\mathcal{V} = \{v_1, v_2, \ldots, v_L\}$ with $L$ frames, we perform $K$ sampling iterations to ensure robust temporal coverage.
For the $k$-th iteration ($k \in \{1, \ldots, K\}$), we sample $N$ frames at low resolution using uniformly spaced intervals with a shifted starting offset:
\begin{equation}
    \mathcal{I}^{(k)} = \{v_{j_1^{(k)}}, v_{j_2^{(k)}}, \ldots, v_{j_N^{(k)}}\},
    \quad j_i^{(k)} = \left\lfloor \frac{(i-1) \cdot L}{N} + \frac{k \cdot L}{K \cdot N} \right\rfloor.
\end{equation}
This offset-based sampling ensures different iterations capture complementary temporal slices of the video.
We insert frame index tags \texttt{<frame i>:} before each image \texttt{i} to enable the MLLM to reference specific frames in its response.
Each sampled set $\mathcal{I}^{(k)}$ is fed to the MLLM with a specialized designed prompt \jynew{(shown in the Appendix Tab.~\ref{prompt_stage1})}, which enables the MLLM to jointly perform semantic reasoning (identifying the functional object), temporal reasoning (selecting which frame best shows it), and spatial grounding (localizing the part), all in a single forward pass.
The model outputs a structured response indicating the functional object name, selected frame index, and point coordinates:
\begin{equation}
    (o_{\text{func}}^{(k)},\, f^{(k)},\, \mathbf{P}^{(k)}) = \Phi_{\text{R1}}(T,\, \mathcal{I}^{(k)}),
\end{equation}
where $o_{\text{func}}^{(k)}$ is the predicted functional object name, $f^{(k)}$ is the selected frame index, and $\mathbf{P}^{(k)} = \{(x_i, y_i)\}$ is the set of predicted points.
To ensure robust temporal coverage, all $K$ predictions for which a valid frame index is returned are retained and collectively forwarded to the fine stage.
This active multi-sampling ensemble prevents any single temporal slice from dominating: diverse offsets collectively cover the full video, and the union of candidate frames from all iterations provides Round 2 with a rich set of starting points for high-resolution refinement.

\noindent\textbf{Fine stage (Round 2): Native zoom-in via dense temporal window at high resolution.}
The coarse stage produces a set of candidate frames $\mathcal{F} = \{f^{(k)} \mid r_k = \text{valid}\}_{k=1}^{K}$, each from a different temporal offset but all operating at reduced resolution.
Mimicking human-like zoom-in perception, for each candidate frame $f^{(k)} \in \mathcal{F}$ the fine stage extracts a dense temporal window $\mathcal{W}^{(k)}$ around it from the full video and processes it at native high resolution:
\begin{equation}
    \mathcal{W}^{(k)} = \{v_{f^{(k)}-\Delta}, \ldots, v_{f^{(k)}}, \ldots, v_{f^{(k)}+\Delta}\},
\end{equation}
where $\Delta$ defines the window radius.
The temporal window is calculated based on the sampling interval: if Round 1 samples $N$ frames from a video of duration $t_{\text{total}}$, each interval spans $\Delta t = t_{\text{total}} / (N-1)$, and we extract frames within $\pm \Delta t / 2$ seconds of the candidate frame's timestamp.
For each frame $I_j \in \mathcal{W}^{(k)}$, we query the MLLM independently with a high resolution image \jynew{(as detailed in the Appendix Tab.~\ref{prompt_stage2})}:
This enables fine-grained localization for each frame independently.
Processing all frames in $\mathcal{W}^{(k)}$ provides multiple candidate predictions per window:
\begin{equation}
    \{\mathbf{P}_j\}_{j \in \mathcal{W}^{(k)}} = \{\Phi_{\text{R2}}(T,\, o_{\text{func}}^j,\, I_j)\}_{j \in \mathcal{W}^{(k)}},
\end{equation}
where $o_{\text{func}}^j$ is the functional object predicted by the coarse stage.
This is repeated for all $K$ candidate windows, yielding a combined set of per-frame point predictions across all temporal windows.

The fine stage addresses all three failure modes of passive frame selection by providing dense multi-view context from multiple angles and timestamps at high resolution:
(1) Missing spatial context becomes visible across the temporal window as multiple viewpoints collectively cover the full scene context.
(2) Partially visible targets gain complete coverage through multi-view aggregation, resolving spatial ambiguities that are unresolvable in any single isolated frame.
(3) The resolution transition from the coarse stage to native high resolution delivers zoom-in capability for small functional parts without any image cropping: because the full frame is retained rather than cropped, the surrounding context remains available for disambiguation throughout.
Crucially, if the coarse stage's initial estimate is slightly off-target, the fine stage can self-correct by reasoning over the complete high-resolution scene, a recovery that is impossible in a crop-and-reprocess agent once the wrong region has been committed to.
We aggregate per-frame predictions from all $K$ temporal windows into a combined set $\{\mathbf{P}_j\}$ that leverages holistic spatial-temporal understanding.

\subsection{Visual Mask Generation and Verification}
\label{sec:segmentation}

From the fine stage, we obtain per-frame point predictions $\{\mathbf{P}_j\}$ aggregated across all $K$ temporal windows, where each $\mathbf{P}_j = \{(x_i^j, y_i^j)\}_{i=1}^{M_j}$ contains $M_j$ points. We use them as point prompts for SAM3~\cite{sam3}.
For each point prompt $p_i^j = (x_i^j, y_i^j)$ in frame $I_j$ across all $K$ windows, it generates a binary mask $M_i^j$.
As SAM3 may occasionally over-segment by including parent objects, we apply MLLM-based visual mask verification as the final stage of the unified reasoning framework.
Rather than verifying points through text-level queries (as in TASA~\cite{tasa}), we render each mask as a colored overlay on its source frame and present it directly to the MLLM for visual judgment, enabling the model to reason about whether the highlighted region is precisely the intended functional part.
We create an overlay image $\tilde{I}_j = \text{Overlay}(I_j, M_i^j, \alpha)$ by highlighting the masked region with transparency $\alpha$.
The verification is formulated as:
\begin{equation}
    b_i^j = \Phi_{\text{verify}}(\tilde{I}_j, o_{\text{func}}^*) \in \{\text{YES}, \text{NO}\}.
\end{equation}
The model verifies two criteria:
(1) The highlighted region is the target functional object.
(2) It does not include parent or containing objects. \jynew{Detailed process is described in the Appendix.}
Only masks receiving $b_i^j = \text{YES}$ are retained for 3D lifting $\mathcal{M}_{\text{verified}}$.
By reusing the same MLLM backbone that performed Stage 1, the verification maintains semantic consistency throughout the unified reasoning pipeline and substantially improves precision without sacrificing recall.

\subsection{Multi-view Agreement and 3D Lifting}
\label{sec:lifting}

To obtain the final 3D masks, following~\cite{Functionality}, we lift the verified 2D masks from all processed frames onto the point cloud using camera poses.
We employ multi-view agreement to filter spurious predictions:
For each 3D point $c_i \in \mathcal{C}$, we count how many 2D masks project onto it:
\begin{equation}
    s_i = \sum_{k=1}^K \lvert \{p^k \mid \Gamma^{k}(p^k) = c_i, p^k \in M^k \} \rvert,
\end{equation}
where $M^k$ is the 2D functional mask in the $k$-th view, $p^k$ is a pixel in $M^k$, and $\Gamma^{k} \colon \mathbb{Z}^2 \to \mathbb{R}^3$ maps 2D pixels to 3D points via depth and pose.
We normalize $s_i \in [0, 1]$ and threshold at $\tau$ to produce the final mask $\mathcal{M}_F = \{c_i \mid s_i > \tau\}$.
This consensus mechanism suppresses outliers and ensures that only points consistently identified across multiple views are included.

\section{Experiments}
\label{sec:experiments}

\subsection{Experimental Setup}

\noindent\textbf{Dataset.}
We evaluate on SceneFun3D~\cite{delitzas2024scenefun3d}, the only dataset for affordance segmentation in 3D scenes.
It contains 230 high-resolution indoor scene scans split into \textit{split0} (30 scenes, from their validation set) and \textit{split1} (200 scenes, from their training set), following the practice of Fun3DU.

\noindent\textbf{Evaluation metrics.}
Following the original SceneFun3D~\cite{delitzas2024scenefun3d} benchmark design, we report Average Precision at IoU thresholds of 0.25 and 0.5 (AP$_{25}$, AP$_{50}$).
For comprehensive evaluation, we additionally report Average Recall (AR$_{25}$, AR$_{50}$) and mean Intersection-over-Union (mIoU).
These metrics capture both localization accuracy and segmentation quality.

\noindent\textbf{Implementation details.}
We implement two variants (8B, 30B) of UniFunc3D using MLLM models in different sizes of Qwen3-VL~\cite{qwen3vl}.
For Round 1, we sample $N=64$ frames per video at $512 \times 384$ resolution with $K=4$ sampling iterations using different temporal offsets.
For Round 2, we extract temporal windows of size $2\Delta+1$ frames around the selected frame.
We use SAM3~\cite{sam3} for mask generation with a confidence threshold of 0.5. The overlay transparent parameter $\alpha$ is set to 0.7.
The final 3D mask is obtained by thresholding the multi-view agreement heatmap at $\tau=0.7$.
All models are frozen and used without fine-tuning, making UniFunc3D fully training-free.

\noindent\textbf{Baselines.}
We compare against both training-free and training-based methods:
(1) OV-3D segmentation methods \textbf{OpenMask3D}~\cite{takmaz2024openmask3d}, \textbf{LERF}~\cite{kerr2023lerf}, and \textbf{OpenIns3D}~\cite{huang2024openins3d}, which lack task-specific reasoning.
(2) Training-based methods \textbf{TASA}~\cite{tasa} and \textbf{AffordBot}~\cite{affordbot}, which utilize Qwen2.5-VL-72B and require training on 3D labels.
(3) \textbf{Fun3DU}~\cite{Functionality}, the pioneering training-free method.

\subsection{Main Results}

\begin{table}[t]
\newcommand{\gain}[1]{{\scriptsize #1}}
\centering
\caption{\textbf{Results on SceneFun3D.} UniFunc3D achieves the best performance across all metrics on both split0 and split1, outperforming both training-free and training-based methods. Percentages in parentheses denote relative improvements over Fun3DU.}
\label{tab:main}
\small
\resizebox{\textwidth}{!}{
\begin{tabular}{l|ccccc|ccccc}
\toprule
 & \multicolumn{5}{c|}{Split0 (30 scenes, val)} & \multicolumn{5}{c}{Split1 (200 scenes, train)} \\
\cmidrule(lr){2-6}\cmidrule(lr){7-11}
Method & AP$_{50}\uparrow$ & AP$_{25}\uparrow$ & AR$_{50}\uparrow$ & AR$_{25}\uparrow$ & mIoU$\uparrow$ & AP$_{50}\uparrow$ & AP$_{25}\uparrow$ & AR$_{50}\uparrow$ & AR$_{25}\uparrow$ & mIoU$\uparrow$ \\
\midrule
\multicolumn{11}{l}{\textit{Open-vocabulary 3D segmentation methods:}}\\\midrule
OpenMask3D~\cite{takmaz2024openmask3d} & 0.2 & 0.4 & 24.5 & 27.0 & 0.2 & 0.0 & 0.0 & 1.4 & 2.6 & 0.1 \\
LERF~\cite{kerr2023lerf} & 0.0 & 0.0 & 35.1 & 36.0 & 0.0 & 0.0 & 0.0 & 24.6 & 25.1 & 0.0 \\
OpenIns3D~\cite{huang2024openins3d} & 0.0 & 0.0 & 46.7 & 51.5 & 0.1 & 0.0 & 0.0 & 37.1 & 39.9 & 0.1 \\
\midrule\midrule
\multicolumn{11}{l}{\textit{Training-based 3D affordance segmentation methods:}}\\\midrule
TASA-72B~\cite{tasa} & 26.9 & 28.6 & -- & -- & 19.7 & \multicolumn{5}{c}{\textit{trained on split1; not evaluated}} \\
AffordBot-72B~\cite{affordbot} & 20.91 & 24.76 & 18.99 & 22.84 & 14.42 & \multicolumn{5}{c}{\textit{trained on split1; not evaluated}} \\
\midrule\midrule
\multicolumn{11}{l}{\textit{Training-free 3D affordance segmentation methods:}}\\\midrule
\makecell[l]{Fun3DU$^\dagger$~\cite{Functionality}} & 16.9 & 33.3 & 38.2 & 46.7 & 15.2 & 12.6 & 23.1 & 32.9 & 40.5 & 11.5 \\ [1ex]
\multirow{2}{*}{\makecell[c]{UniFunc3D-8B}} & 23.82 & 44.04 & 46.07 & 55.51 & 20.92 & 16.24 & 29.02 & 38.91 & 48.15 & 14.23 \\ 
 & \gain{(+40.9\%)} & \gain{(+32.3\%)} & \gain{(+20.6\%)} & \gain{(+18.9\%)} & \gain{(+37.6\%)} & \gain{(+28.9\%)} & \gain{(+25.6\%)} & \gain{(+18.3\%)} & \gain{(+18.9\%)} & \gain{(+23.7\%)} \\ [1ex]
\multirow{2}{*}{\makecell[c]{UniFunc3D-30B}} & \textbf{31.24} & \textbf{51.01} & \textbf{46.97} & \textbf{58.88} & \textbf{24.30} & \textbf{21.32} & \textbf{35.76} & \textbf{40.03} & \textbf{51.00} & \textbf{17.09} \\
 & \gain{(+84.9\%)} & \gain{(+53.2\%)} & \gain{(+23.0\%)} & \gain{(+26.1\%)} & \gain{(+59.9\%)} & \gain{(+69.2\%)} & \gain{(+54.8\%)} & \gain{(+21.7\%)} & \gain{(+25.9\%)} & \gain{(+48.6\%)} \\
\bottomrule
\end{tabular}}
\\
\raggedright \footnotesize \jynew{$^\dagger$ Fun3DU adopts a sequential offline multi-model pipeline that includes a text-only LLM inference with LLaMa-3.1-9B~\cite{grattafiori2024llama}, contextual object detection with OWLv2~\cite{minderer2023scaling} and RobustSAM~\cite{chen2024robustsam}, Molmo-7B~\cite{molmo} grounding the functional objects on top-ranked views, and SAM~\cite{kirillov2023segany}. Thus, it may be improper to list its ``total'' number of parameters like others in this table.
}
\end{table}

Tab.~\ref{tab:main} presents quantitative results on SceneFun3D.
Our UniFunc3D-30B achieves the best performance across \textbf{all} methods on \textbf{all} reported metrics on \textbf{both} splits.
Remarkably, our training-free approach even outperforms training-based methods, including those using significantly larger models.

\noindent\textbf{Comparison with training-free methods.}
Compared to Fun3DU, the previous state-of-the-art training-free method, UniFunc3D-30B improves AP$_{50}$ by 14.34 points (84.9\% relative gain) and AP$_{25}$ by 17.71 points (53.2\% gain) on split0, showing significantly superior localization accuracy at both tight and loose IoU thresholds.
The mIoU improvement of 9.1 points (59.9\% gain over Fun3DU) further confirms more accurate segmentation of functional parts.
UniFunc3D-8B also demonstrates strong performance, improving AP$_{50}$ by 6.92 points (41.0\% gain) and AP$_{25}$ by 10.74 points (32.3\% gain) over Fun3DU on split0.

\noindent\textbf{Comparison with training-based methods.}
Most notably, UniFunc3D-30B substantially outperforms training-based methods despite being entirely training-free and with a smaller model (30B vs.\ 72B).
Against AffordBot-72B, which requires 1,000 epochs of fine-tuning on SceneFun3D training data, UniFunc3D-30B demonstrates even more striking advantages: AP$_{50}$ (+10.33 points, 49.4\% gain), AP$_{25}$ (+26.25 points, 106.0\% gain), and mIoU (+9.88 points, 68.5\% gain).
These results demonstrate that our unified, visually aware multimodal architecture with temporal grounding is fundamentally more effective than existing specialized training-based approaches.

\noindent\textbf{Open-vocabulary methods.}
Open-vocabulary methods achieve moderate recall but near-zero precision, indicating they over-segment.
They lack the reasoning capability to interpret implicit task descriptions, demonstrating the necessity of task-aware reasoning for 3D affordance segmentation.


\noindent\textbf{Qualitative results.}
Fig.~\ref{fig:qualitative} compares AffordBot, Fun3DU, and our method against ground truth on five representative queries.
Our method clearly outperforms others in solving spatial disambiguation and handling small objects with diverse affordance. For example, for input prompt \texttt{Open the top left drawer of the cabinet with the beauty products on top}, our method can find the correct \texttt{top-left knob}, while AffordBot finds the wrong \texttt{top-right knob} and Fun3DU mistakenly treats the \texttt{drawer} as the functional object.

\newcommand{\rlabel}[1]{\rotatebox{90}{\parbox{2.3cm}{\centering\scriptsize\textbf{#1}}}}
\newcommand{\imgw}{0.185\linewidth}
\newcommand{\qimggt}[2]{\includegraphics[width=\imgw]{selected_results/gt/#1/#2}}
\newcommand{\qimgab}[2]{\includegraphics[width=\imgw]{selected_results/affordbot/#1/#2}}
\newcommand{\qimgfd}[2]{\includegraphics[width=\imgw]{selected_results/fun3du/#1/#2}}
\newcommand{\qimgou}[2]{\includegraphics[width=\imgw]{selected_results/ours/#1/#2}}
\begin{figure}[t]
\centering
\setlength{\tabcolsep}{1pt}
\renewcommand{\arraystretch}{0.5}
\begin{tabular}{c@{\hspace{3pt}}ccccc}

  &
  \parbox[c]{\imgw}{\centering\scriptsize\texttt{Open the top left drawer of the cabinet with the beauty products on top}} &
  \parbox[c]{\imgw}{\centering\scriptsize\texttt{Turn on the ceiling light}} &
  \parbox[c]{\imgw}{\centering\scriptsize\texttt{Control the water flow in the bathhub using the drain control dial}} &
  \parbox[c]{\imgw}{\centering\scriptsize\texttt{Select a washing program}} &
  \parbox[c]{\imgw}{\centering\scriptsize\texttt{Flush the toilet}} \\[4pt]
    \rlabel{AffordBot~\cite{affordbot}} &
    \includegraphics[width=\imgw]{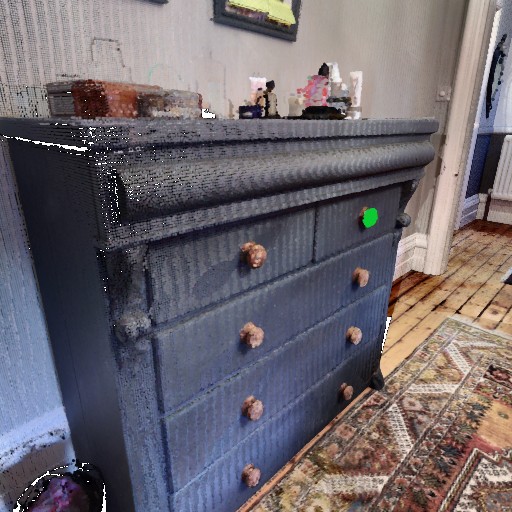} &
    \includegraphics[width=\imgw]{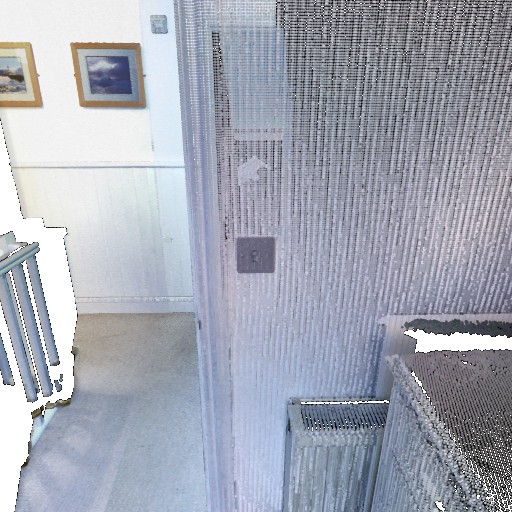} &
    \includegraphics[width=\imgw]{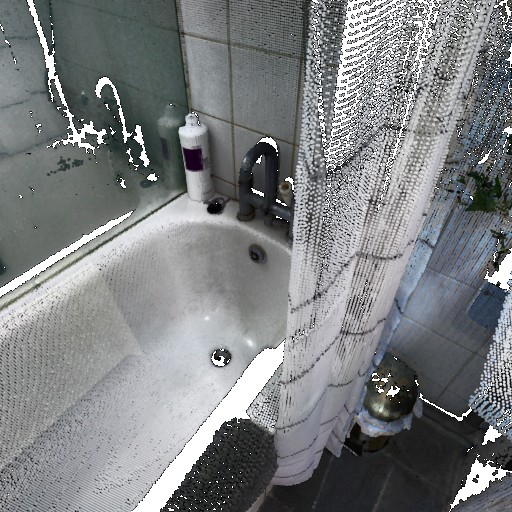} &
    \includegraphics[width=\imgw]{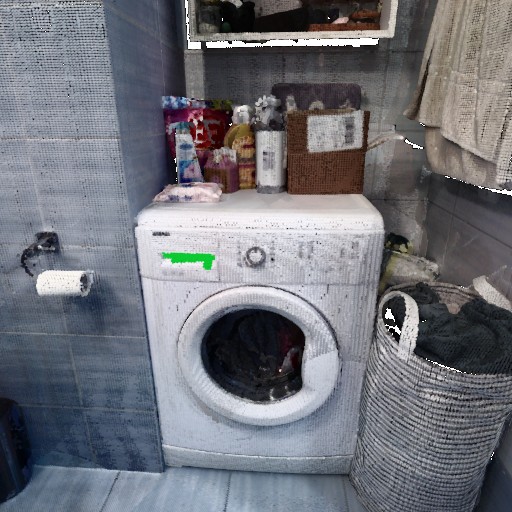} &
    \includegraphics[width=\imgw]{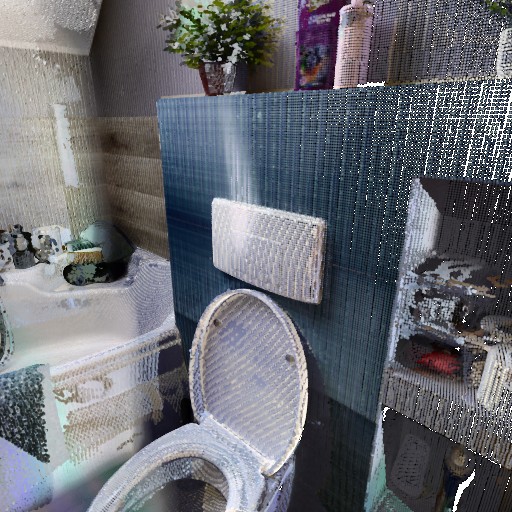} \\[2pt]
  \rlabel{Fun3DU~\cite{Functionality}} &
    \includegraphics[width=\imgw]{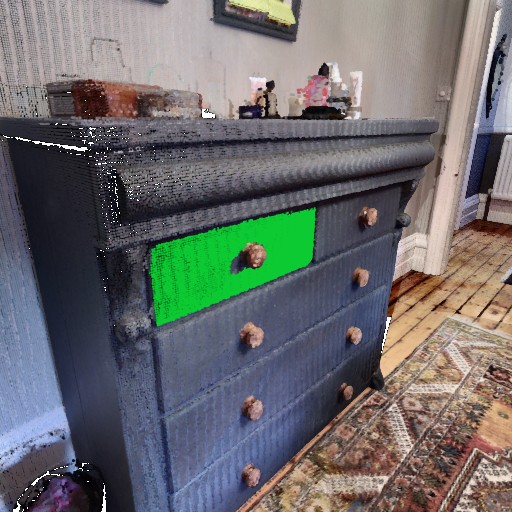} &
    \includegraphics[width=\imgw]{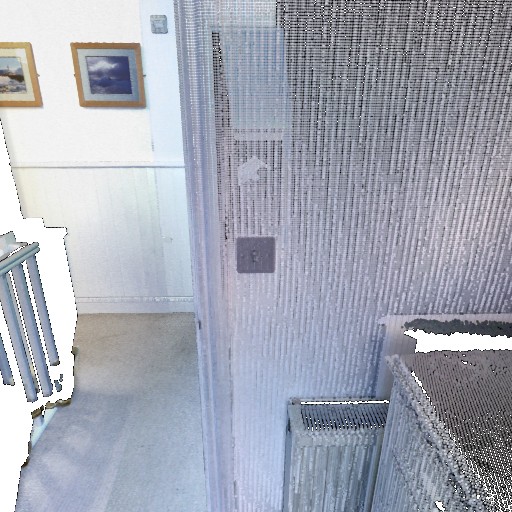} &
    \includegraphics[width=\imgw]{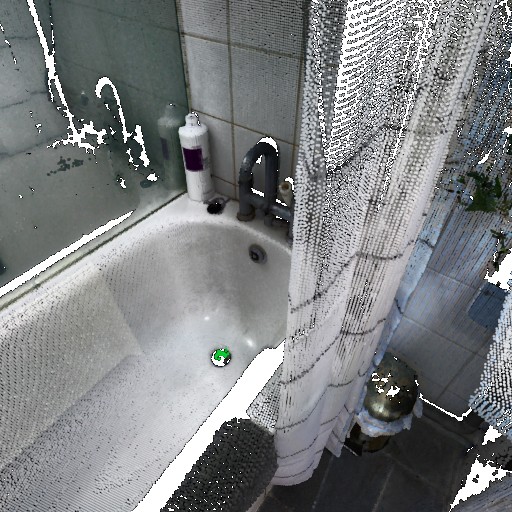} &
    \includegraphics[width=\imgw]{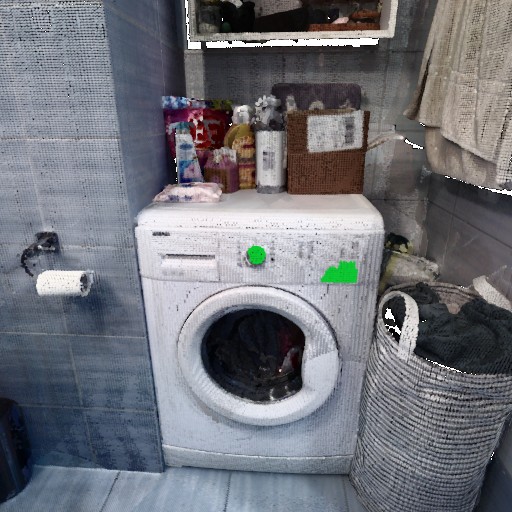} &
    \includegraphics[width=\imgw]{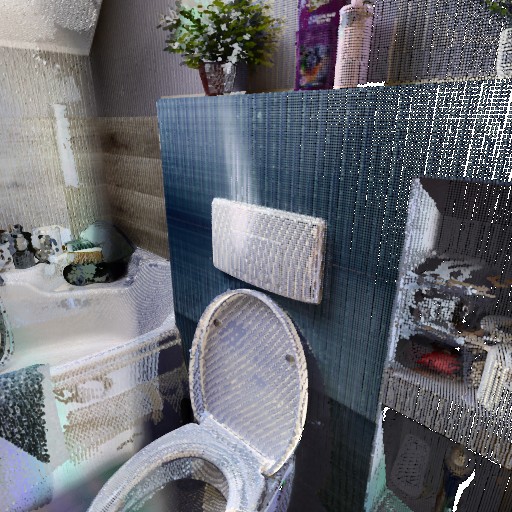} \\[2pt]
  \rlabel{Ours} &
    \includegraphics[width=\imgw]{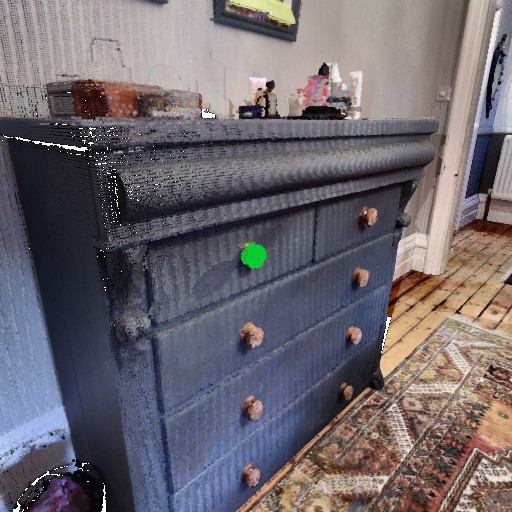} &
    \includegraphics[width=\imgw]{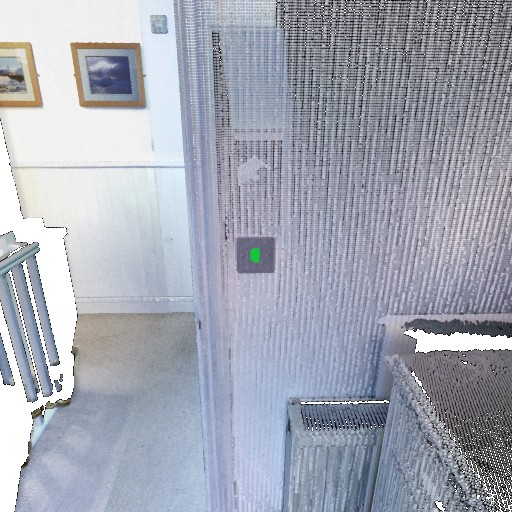} &
    \includegraphics[width=\imgw]{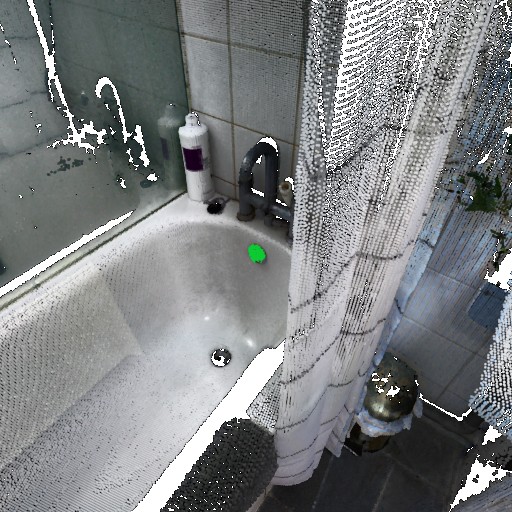} &
    \includegraphics[width=\imgw]{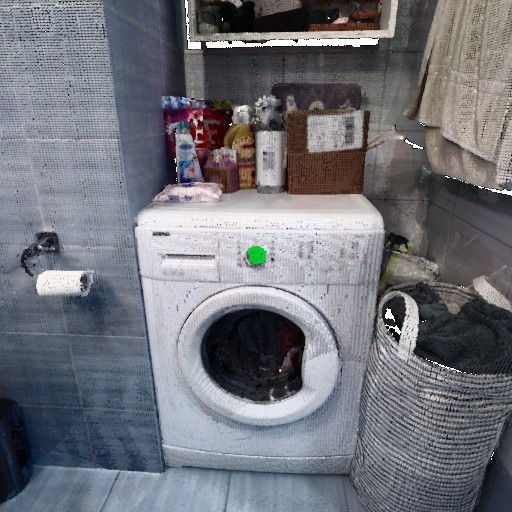} &
    \includegraphics[width=\imgw]{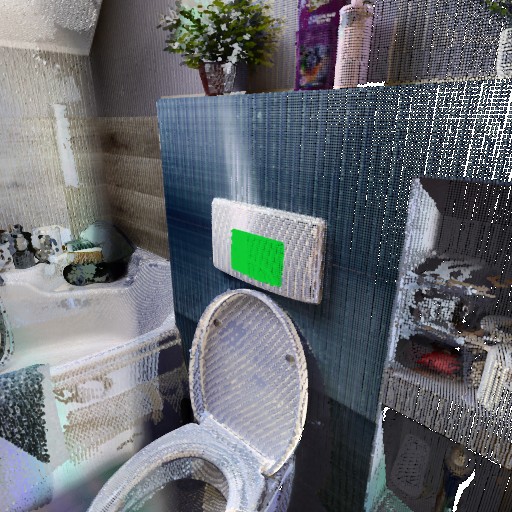} \\[2pt]
  \rlabel{Ground Truth} &
    \includegraphics[width=\imgw]{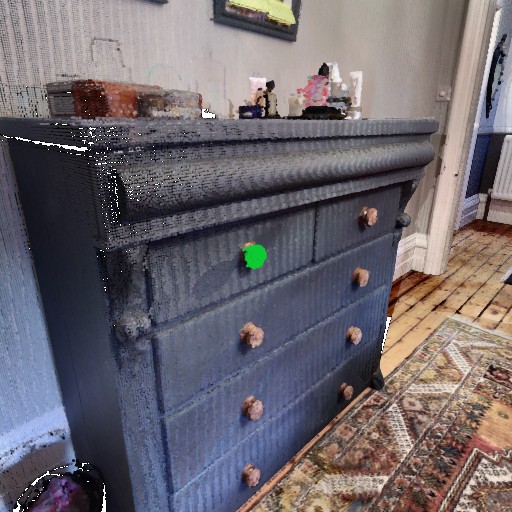} &
    \includegraphics[width=\imgw]{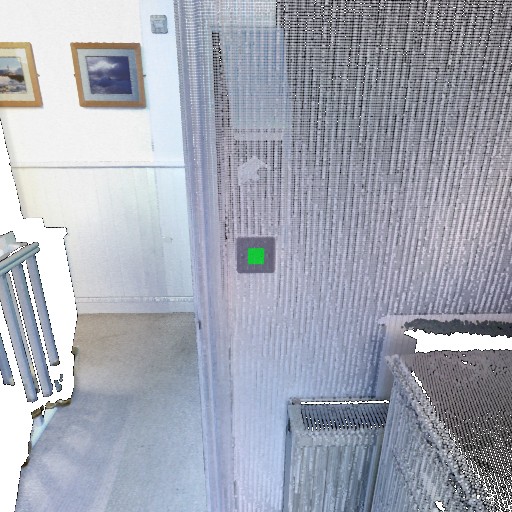} &
    \includegraphics[width=\imgw]{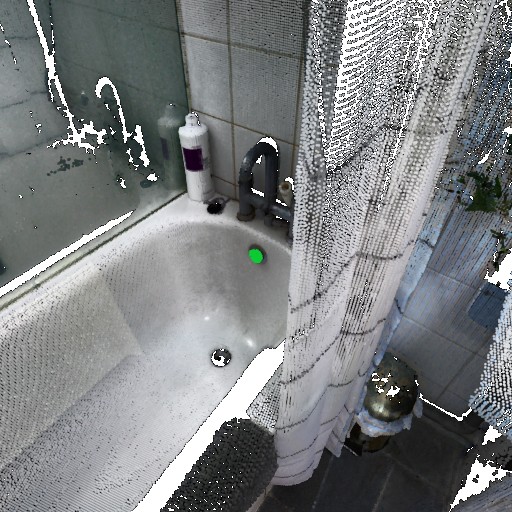} &
    \includegraphics[width=\imgw]{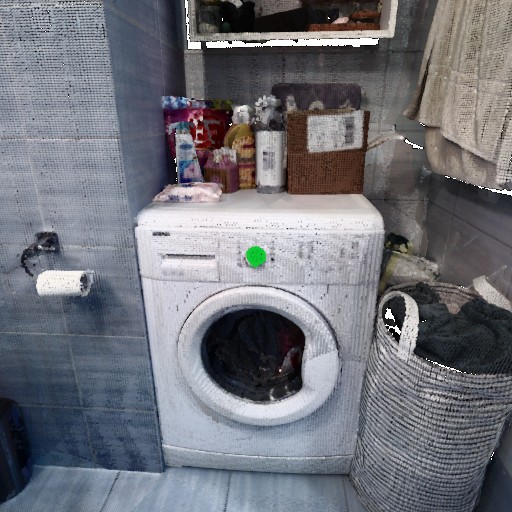} &
    \includegraphics[width=\imgw]{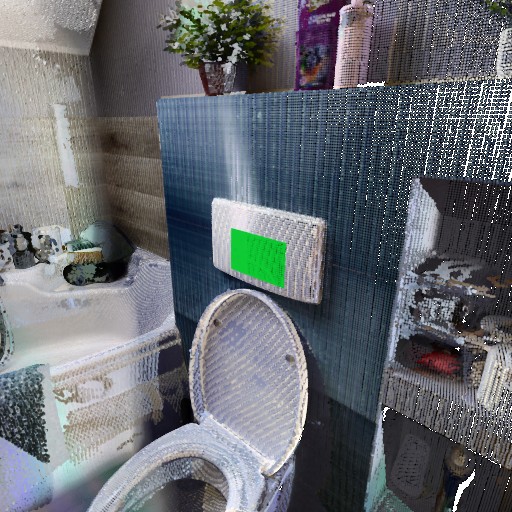} \\[2pt]
\end{tabular}
\vspace{-3mm}
\caption{\textbf{Qualitative comparison.} We show results for representive five queries.}
\label{fig:qualitative}
\end{figure}

\subsection{Ablation Study}

We conduct ablation studies on SceneFun3D split0 using UniFunc3D-8B.

\noindent\textbf{Vanilla MLLMs vs. two-stage grounding.}
Tab.~\ref{tab:ablation_resolution} shows that the vanilla MLLM (\textit{i.e.}, one-stage, directly using Qwen3-VL-8B without other designs except the external segmentation model) approach processing all frames at high resolution without initial coarse selection, severely degrades performance despite using the highest resolution throughout.
This highlights that \textbf{directly using off-the-shelf MLLMs CANNOT solve this problem}: when all frames are processed at high resolution, the model must handle fine-grained detail while lacking the global temporal overview needed to identify the correct frame, and base MLLMs (\textit{i.e.}, Qwen) degrade under such extremely long contexts.
One-stage low-resolution processing is equally limited; although the shorter context better utilizes model capacity, individual frames lack the spatial detail required for precise functional part localization. This matches the practical constraints of using MLLMs.

Our proposed two-stage approach with a single Round 2 frame (low$\rightarrow$high) substantially improves performance, confirming the value of coarse-to-fine processing.
Adding multisampling alone improves further by diversifying Round 1 coverage across shifted temporal offsets and reducing the risk of missing the optimal frame.
Adding temporal window processing alone also yields gains by expanding Round 2 context from a single frame to multiple frames.
Our full model combines both components, achieving the best performance, validating that both components provide complementary benefits.

\begin{table}[t]
\centering
\caption{\textbf{Ablation on high-level designs.} Comparison of one-stage vs.\ two-stage approaches. We also evaluate the impact of \textbf{Multi-Sam.} (Multi-sampling), and \textbf{Temp.} (Temporal window).}
\label{tab:ablation_resolution}
\small
\resizebox{\textwidth}{!}{
\begin{tabular}{l|ccc|cc|cc|c}
\toprule
Strategy & 2-Stage & Multi-Sam. & Temp. & AP$_{50}$ & AP$_{25}$ & AR$_{50}$ & AR$_{25}$ & mIoU \\
\midrule
Qwen3-VL-8B (High-res) & & & & 4.27 & 7.64 & 24.27 & 28.76 & 3.11 \\
Qwen3-VL-8B (Low-res)  & & & & 7.42 & 14.38 & 33.93 & 38.65 & 7.35 \\
\midrule
Low $\rightarrow$ High (Zoom-in) & \checkmark & & & 15.96 & 26.29 & 39.33 & 48.54 & 12.71 \\
Two-stage + Multi-sam.          & \checkmark & \checkmark & & 19.33 & 38.20 & 44.94 & 53.71 & 18.10 \\
Two-stage + Temp. window        & \checkmark & & \checkmark & 17.53 & 35.28 & 41.80 & 51.91 & 17.26 \\
\midrule
\textbf{Ours (Full)}            & \checkmark & \checkmark & \checkmark & \textbf{23.82} & \textbf{44.04} & \textbf{46.07} & \textbf{55.51} & \textbf{20.92} \\
\bottomrule
\end{tabular}
}
\end{table}

\begin{table}[t]
\centering
\caption{\textbf{Ablation on sampling strategies.} Results for UniFunc3D-8B evaluating the impact of sampling iterations ($K$), the temporal window and verification.}
\label{tab:ablation_sampling}
\small
\begin{tabular}{l|cc|cc|cc|c}
\toprule
Configuration & Temp. & Verif. & AP$_{50}$ & AP$_{25}$ & AR$_{50}$ & AR$_{25}$ & mIoU \\
\midrule
$K=1$ (Base) &  & & 15.96 & 26.29 & 39.33 & 48.54 & 12.71 \\
$K=1$, no verify & \checkmark & & 15.06 & 32.58 & 40.90 & 50.11 & 16.00 \\
$K=1$, verify    & \checkmark & \checkmark & 17.53 & 35.28 & 41.80 & 51.91 & 17.26 \\
\midrule
$K=2$, &  & & 15.51 & 28.31 & 39.33 & 46.07 & 14.40 \\
$K=2$, no window &  & \checkmark & 12.58 & 27.42 & 40.00 & 47.42 & 13.87 \\
$K=2$, no verify & \checkmark & & 21.12 & 39.10 & \textbf{46.97} & 56.18 & 19.57 \\
$K=2$, verify    & \checkmark & \checkmark & 22.02 & 43.60 & 46.29 & 54.38 & 20.69 \\
\midrule
$K=4$, &  & & 15.96 & 30.11 & 41.57 & 50.11 & 15.30 \\
$K=4$, no window &  & \checkmark & 19.33 & 38.20 & 44.94 & 53.71 & 18.10 \\
$K=4$, no verify & \checkmark & & 21.35 & 41.12 & 46.74 & 55.51 & 19.99 \\
$K=4$, verify    & \checkmark & \checkmark & \textbf{23.82} & \textbf{44.04} & 46.07 & 55.51 & \textbf{20.92} \\
\midrule
$K=8$, no window &  & \checkmark & 22.47 & 42.70 & 42.70 & 52.81 & 19.48 \\
$K=8$, no verify & \checkmark & & 21.57 & 42.25 & 46.52 & \textbf{56.40} & 19.95 \\
\bottomrule
\end{tabular}
\end{table}

\noindent\textbf{Effect of temporal window.}
The temporal window provides the largest single gain.
Adding the temporal window when $K=2$ raises AP$_{50}$ to 21.12 (+5.61) and AP$_{25}$ from 28.31 to 39.10 (+10.79).
The same trend holds at $K=4$.
The temporal window also critically enables verification to function effectively: applying verification without the window at $K=2$ slightly \textit{hurts} performance, because there are too few candidate frames for reliable confidence ranking.

\begin{figure}[t]
    \centering
    \begin{tabular}{ccc}
        \includegraphics[width=0.26\linewidth]{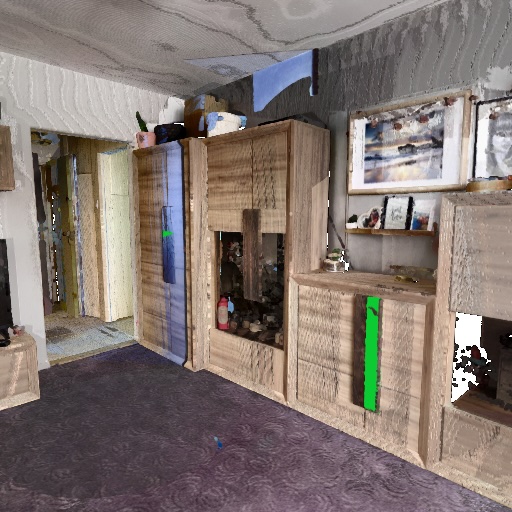} &
        \includegraphics[width=0.26\linewidth]{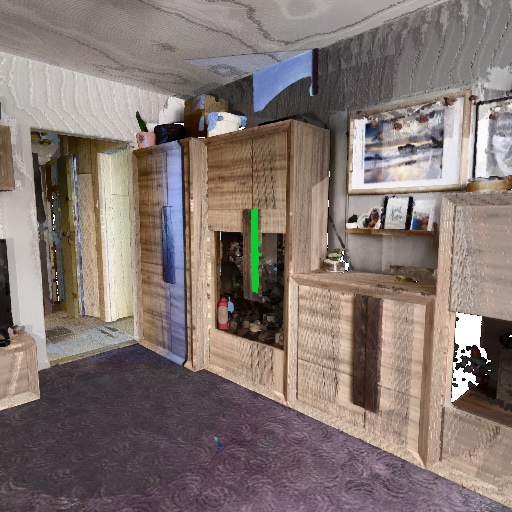} &
        \includegraphics[width=0.26\linewidth]{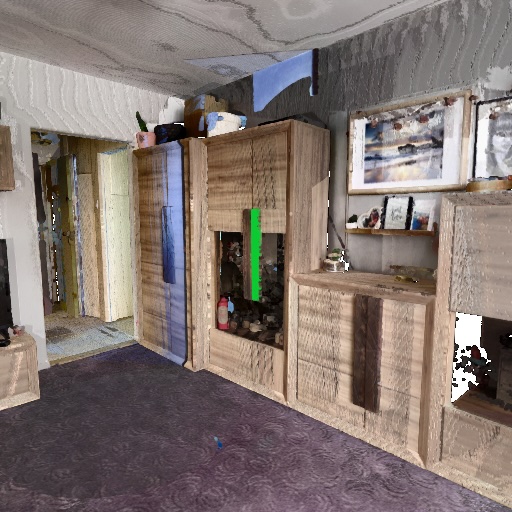} \\
        \footnotesize (a) Ablation ($K=1$) & \footnotesize (b) UniFunc3D (Ours) & \footnotesize (c) Ground Truth
    \end{tabular}
    \caption{Visual comparison. Input prompt: \texttt{open the right door of the wooden display cabinet to the left of the paintings}.}
    \label{fig:visual}
\end{figure}

\noindent\textbf{Effect of sampling iterations ($K$).}
With the temporal window enabled, increasing from $K=1$ to $K=2$ provides substantial gains, as multiple shifted offsets reduce the risk of missing the optimal frame region.
Further increasing to $K=4$ yields only marginal improvement over $K=2$ without verification (21.35 vs.\ 21.12 AP$_{50}$), and $K=8$ shows similarly diminishing returns (21.57 AP$_{50}$), suggesting $K=4$ is the practical saturation point. Fig.~\ref{fig:visual} shows that single-iteration sampling ($K=1$) fails due to limited views. Without diverse temporal candidates, the model may produce spatially inconsistent predictions, such as misidentifying an adjacent handle.
Our multi-sampling strategy resolves these local ambiguities by providing broader visual evidence.

\noindent\textbf{Effect of verification.}
Verification is most effective when the candidate pool is large enough.
With $K=2$, it slightly improves AP$_{50}$ (22.02 vs.\ 21.12) and substantially improves AP$_{25}$ (43.60 vs.\ 39.10).
With $K=4$, the gains are more consistent: +2.47 AP$_{50}$, +2.92 AP$_{25}$, and +0.93 mIoU, as the larger pool gives the filter more room to select high-confidence frames.

\noindent\textbf{Effect of base models.} To evaluate the robustness of our approach, we conduct an ablation experiment on the base MLLM models we used. For a fair comparison, we adopt Qwen2.5-VL-72B~\cite{qwen2.5-VL} as our backbone, aligning with the architecture used by contemporary training-based baselines, specifically TASA~\cite{tasa} and AffordBot~\cite{affordbot}. As shown in Tab.~\ref{tab:ablation_base}, the experimental results demonstrate that our training-free model, when integrated with Qwen2.5-VL-72B, consistently outperforms existing training-based methods across nearly all evaluation protocols.

Specifically, we surpass TASA in $2$ out of $3$ metrics, including $AP_{50}$, $AP_{25}$, and mIoU, and exceed AffordBot's performance across $AP_{50}$, $AP_{25}$, $AR_{50}$, $AR_{25}$, and mIoU. While TASA maintains a marginal lead in $AP_{50}$ in one specific configuration, our model's superiority across diverse metrics underscores the efficacy of our proposed pipeline enhancements. This shows the generalizability of our framework, proving that its performance gains are not merely a byproduct of a superior base model but stem from our intrinsic methodological design.

\begin{table}[t]
\centering
\caption{\textbf{Ablation on base models.} Best-performed metrics with Qwen2.5-VL-72B (the top group) are marked in \textbf{bold}. We also provide our models with Qwen3-VL for reference.}
\label{tab:ablation_base}
\small
\begin{tabular}{l|cc|cc|c}
\toprule
Baselines & AP$_{50}$ & AP$_{25}$ & AR$_{50}$ & AR$_{25}$ & mIoU \\
\midrule
TASA-72B$^\dagger$~\cite{tasa} & \textbf{26.9} & 28.6 & -- & -- & 19.7 \\
AffordBot-72B$^\dagger$~\cite{affordbot} & 20.91 & 24.76 & 18.99 & 22.84 & 14.42  \\
Ours (Qwen2.5-VL-72B)   & 22.02 & \textbf{40.90}  & \textbf{48.09} & \textbf{59.78} & \textbf{20.07} \\
\midrule
Ours (Qwen3VL-8B)   & 23.82 & 44.04 & 46.07 & 55.51 & 20.92 \\
Ours (Qwen3VL-30B)  & 31.24 & 51.01 & 46.97 & 58.88 & 24.30 \\
\bottomrule
\end{tabular}
\\
\raggedright \footnotesize $^\dagger$ Qwen2.5-VL-72B is used according to the authors.
\end{table}

\noindent\textbf{Effect of verification models.}
To verify the design of our verification models, we change the segmentation model we used (\textit{i.e.}, SAM3) to the previous SAM~\cite{kirillov2023segany}, used by Fun3DU~\cite{Functionality}. As shown in Tab.~\ref{tab:ablation_verification}, We do not observe significant improvement by changing the segmentation model. This indicates that the most improvement of our UniFunc3D does \textbf{NOT} come from the modern or powerful segmentor but the overall pipeline design.


\begin{table}[t]
\centering
\small

\begin{minipage}[t]{0.48\textwidth}
\centering
\caption{\textbf{Ablation on verification models.}}
\label{tab:ablation_verification}
\resizebox{\linewidth}{!}{%
\begin{tabular}{l|cc|cc|c}
\toprule
Baselines & AP$_{50}$ & AP$_{25}$ & AR$_{50}$ & AR$_{25}$ & mIoU \\
\midrule
SAM~\cite{kirillov2023segany}
& 23.56 & \textbf{44.23} & 45.91 & 54.99 & 20.78 \\
SAM3~\cite{sam3}
& \textbf{23.82} & 44.04 & \textbf{46.07}
& \textbf{55.51} & \textbf{20.92} \\
\bottomrule
\end{tabular}%
}
\end{minipage}
\hfill
\begin{minipage}[t]{0.48\textwidth}
\centering
\caption{\textbf{Ablation on radius settings.}}
\label{tab:ablation_radius}
\resizebox{\linewidth}{!}{%
\begin{tabular}{l|cc|cc|c}
\toprule
Radius & AP$_{50}$ & AP$_{25}$ & AR$_{50}$ & AR$_{25}$ & mIoU \\
\midrule
$0.5\Delta$
& 22.92 & \textbf{44.94} & 45.84 & 55.28 & 20.80 \\
$1.0\Delta$
& 23.82 & 44.04 & \textbf{46.07} & 55.51 & 20.92 \\
$1.5\Delta$
& \textbf{24.49} & 44.27 & 45.62
& \textbf{55.96} & \textbf{20.94} \\
\bottomrule
\end{tabular}%
}
\end{minipage}

\end{table}

\noindent\textbf{Effect of hyperparameter $\Delta$.}
Table~\ref{tab:ablation_radius} shows the effect of hyperparameter $\Delta$, which is relatively insensitive within this range, whereas the number of shifted sampling iterations $K$ has a larger effect.


\section{Conclusion}
\label{sec:conclusion}

We presented UniFunc3D, a unified framework for 3D scene affordance segmentation that consolidates semantic reasoning, temporal grounding, and spatial localization into a single multimodal large language model.
By processing video frames with our active spatial-temporal grounding process, UniFunc3D eliminates the visual blindness and cascading errors inherent in fragmented multi-model pipelines.
Extensive experiments show that UniFunc3D achieves state-of-the-art performance across all metrics, surpassing both training-free and training-based methods by a large margin.

\begin{ack}

The research is supported in part by the Early Career Scheme of the Research
Grants Council (RGC) of the Hong Kong SAR under grant No.~26202321, ITF
PRP/046/24FX, Science \& Technology Cooperation Program of Shandong under
grant No. SDST26EG01, SAIL Research Project, and HKUST-Zeekr Collaborative Research Fund. 

\end{ack}

\clearpage
\bibliographystyle{abbrv}
\bibliography{main}


\clearpage
\appendix

\section{Additional Experiments and Discussions}

\subsection{Efficiency Analysis}
\label{sec:eff}
We conduct an efficiency analysis with training-free methods Fun3DU~\cite{Functionality} and our UniFunc3D.
We omit metrics such as total parameters or FLOPs, as the multi-model design of Fun3DU makes a unified estimation of these values inherently imprecise.
Furthermore, the standard FLOP metric fails to account for critical non-neural operations essential to the Fun3DU pipeline, such as 3D lifting and point cloud aggregation. These geometric transformations contribute significantly to the overall execution latency but are not reflected in traditional neural network complexity measures, making end-to-end inference time a more representative benchmark for efficiency.
Thus, we adopt Total Inference Latency (\textit{i.e.}, time per scene and time per query), a simple but better metric, for comparing such heterogeneous architectures based on the same platform.
Besides, we exclude training-based methods (e.g., TASA~\cite{tasa} and AffordBot~\cite{affordbot}) from this specific analysis. This is due to the partial unavailability of their source code, specifically the training code for AffordBot's 3D refinement, making a reproducible comparison impossible in this regard.

\begin{table}[h]
\centering
\caption{\textbf{Time comparison.} UniFunc3D-8B achieves over 5.2 times speedup compared to Fun3DU while maintaining superior accuracy.}
\label{tab:efficiency}
\small
\begin{tabular}{l|c|c|c}
\toprule
Method & Time per scene & Time per query & Speedup \\
\midrule
Fun3DU~\cite{Functionality} & $\sim$82 min & $\sim$332 sec & 1$\times$ \\
\textbf{UniFunc3D-8B } & $\sim$16 min & $\sim$53 sec & \textbf{5.2$\times$} \\
\textbf{UniFunc3D-30B} & $\sim$26 min & $\sim$99 sec & \textbf{3.2$\times$} \\
\bottomrule
\end{tabular}
\end{table}

Tab.~\ref{tab:efficiency} compares the total processing time averaged over the validation split (30 scenes) on the same platform with eight H100s.
Fun3DU requires approximately 82 minutes per scene due to its multi-stage pipeline using different models: text-only LLM inference with LLaMa-3.1-9B~\cite{grattafiori2024llama}, contextual object detection with OWLv2~\cite{minderer2023scaling} and RobustSAM~\cite{chen2024robustsam} on hundreds of input video frames, Molmo-7B~\cite{molmo} grounding the functional objects on top-ranked views, and then do functional object segmentation by SAM~\cite{kirillov2023segany}.
UniFunc3D reduces this to just 26 minutes per scene, a ${\sim}$3.2 times speedup, by consolidating reasoning and perception into a single MLLM and using the two-round temporal strategy to avoid exhaustive frame processing.
\jynew{Another benefit from our single MLLM design is that it can better utilize the advantages of some modern efficient inference and serving engines for LLMs such as VLLM~\cite{vllm} with PagedAttention and KV-cache. }
This efficiency gain is achieved while simultaneously improving accuracy, demonstrating the advantages of our unified architecture.

\subsection{Experimental Details}
\label{sec:exp}
All experiments are conducted on a high-performance computing node equipped with an Intel Xeon Platinum 8468 (2.1GHz, 48-core) CPU and eight NVIDIA H100 GPUs. To ensure a rigorous and controlled environment, all baselines and our proposed model are evaluated on this same hardware platform. \jynew{It is worth noting that our method can still run on a single H100 but we do not have enough time to re-run all baseline and ablation experiments on a different platform, thus leaving the experiments to the community when the code is available in the near future.}

For the baseline results presented in the main paper, we follow a best-of-both strategy: we report the higher performance between the values provided in the original papers and our own reproduced if their source code is available. For instance, as shown in Tab.~\ref{tab:reimplementation}, our reproduced results of AffordBot~\cite{affordbot} achieved slightly higher performance than the results reported in the original work.
For TASA~\cite{tasa} and Fun3DU~\cite{Functionality}, the reported results are better than our reproduced results.
Consequently, we adopt these improved metrics for our main comparisons to ensure that our performance gains are measured against the strongest possible version of the state-of-the-art.
This conservative reporting approach ensures that the observed improvements of UniFunc3D are not attributed to suboptimal baseline configurations but rather to our architectural innovations.

Table~\ref{prompt_stage1} and Table~\ref{prompt_stage2} show the system prompt used in the coarse stage and the fine stage of our proposed method, respectively.

\begin{table}[t]
\centering
\caption{\textbf{Comparative results of different implementations of baseline methods.}}
\label{tab:reimplementation}
\small
\begin{tabular}{l|cc|cc|c}
\toprule
Implementation & AP$_{50}$ & AP$_{25}$ & AR$_{50}$ & AR$_{25}$ & mIoU \\
\midrule
TASA-72B (Reported) & \textbf{26.9} & \textbf{28.6} & -- & -- & \textbf{19.7} \\
TASA-72B (Our Reproduced) & 19.72 & 23.58  & 21.17 & 25.21 & 14.61 \\
\midrule
AffordBot-72B (Reported)~\cite{affordbot} & 20.00 & 23.30 & - & - & 14.00 \\
AffordBot-72B (Our Reproduced) & \textbf{20.91} & \textbf{24.76} & \textbf{18.99} & \textbf{22.84} & \textbf{14.42} \\
\midrule
Fun3DU (Reported)~\cite{Functionality} & \textbf{16.9} & \textbf{33.3} & \textbf{38.2} & \textbf{46.7} & \textbf{15.2} \\
Fun3DU (Our Reproduced) & 12.13 & 23.15 & 33.93 & 40.90 & 11.34 \\
\bottomrule
\end{tabular}
\end{table}

\begin{table}[h]
\begin{minipage}{1.0\columnwidth}\vspace{0mm}    \centering
\begin{tcolorbox}
\texttt{Given these frames and the task: [$T$], complete three tasks:}\\
\texttt{1. Identify the functional object needed to accomplish the task.}\\
\texttt{2. Select the key frame that best shows the functional object.}\\
\texttt{3. Identify a single affordance point (x, y) on the functional object.}\\
\texttt{Output format: <affordance> functional object: ...; <frame n>: ...; (x, y) </affordance>}
\end{tcolorbox}
\end{minipage}
\caption{Prompt used in the coarse grounding stage of our UniFunc3D. T is the input task prompt.}
\label{prompt_stage1}
\end{table}

\begin{table}[h]
\begin{minipage}{1.0\columnwidth}\vspace{0mm}    \centering
\begin{tcolorbox}
\texttt{``Identify the affordance point on the [$o_{\text{func}}^*$] in order to [$T$].}\\
\texttt{Output format: <affordance> (x, y) </affordance>.''}
\end{tcolorbox}
\end{minipage}
\caption{Prompt used in the fine grounding stage of our UniFunc3D. T is the input task prompt. $o_{\text{func}}^j$ is the functional object predicted by the coarse stage.}
\label{prompt_stage2}
\end{table}

\begin{table}[h]
\begin{minipage}{1.0\columnwidth}\vspace{0mm}    \centering
\begin{tcolorbox}
\texttt{Does the RED highlighted region show ONLY [$o_{\text{func}}^*$]?}
\end{tcolorbox}
\end{minipage}
\caption{Prompt used in the verification stage of our UniFunc3D. $o_{\text{func}}^j$ is the functional object predicted by the previous stage.}
\label{prompt_verify}
\end{table}

\subsection{Limitations and Future Work}
\label{sec:limit}
While UniFunc3D significantly advances the state-of-the-art, challenges remain for extremely small functional parts ($<$0.1\% of image area) or scenes with severe occlusion.
Future work could explore different zoom-in mechanisms or incorporate explicit 3D geometric reasoning directly within the MLLM to handle complex spatial layouts.
Additionally, extending UniFunc3D to support interactive refinement in which users provide feedback to correct predictions would be valuable for real-world robotic applications.

\subsection{Additional Visual Comparisons for Ablation Models}
\label{sec:visual_ablation}
\begin{figure}[t]
    \centering
    \begin{subfigure}[b]{0.32\linewidth}
        \includegraphics[width=\linewidth]{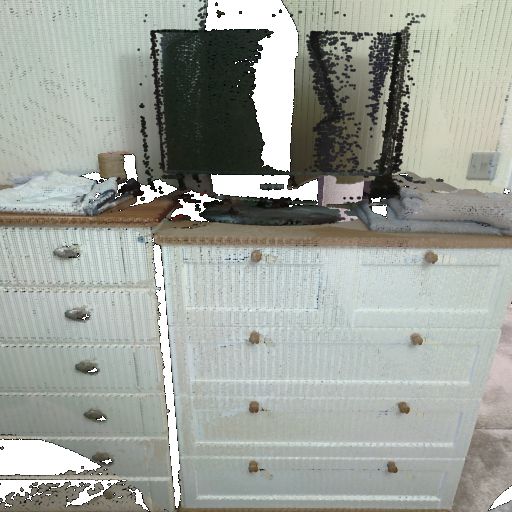}
        \caption*{Base}
    \end{subfigure}\hfill
    \begin{subfigure}[b]{0.32\linewidth}
        \includegraphics[width=\linewidth]{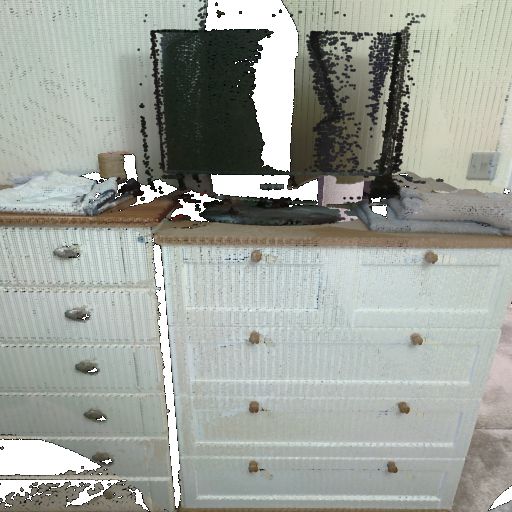}
        \caption*{w/o Multi-view}
    \end{subfigure}\hfill
    \begin{subfigure}[b]{0.32\linewidth}
        \includegraphics[width=\linewidth]{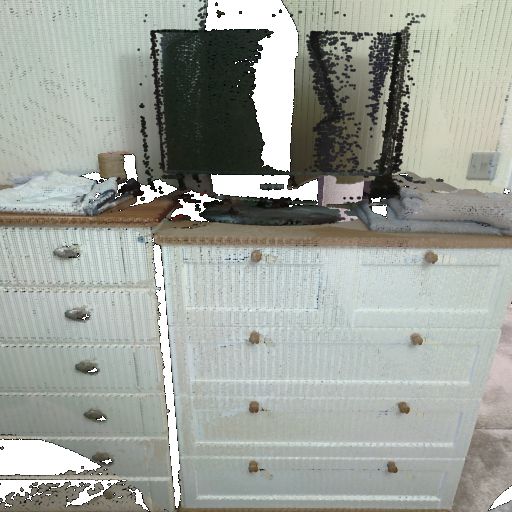}
        \caption*{w/o Temporal}
    \end{subfigure}

    \vspace{0.5em}

    \begin{subfigure}[b]{0.32\linewidth}
        \includegraphics[width=\linewidth]{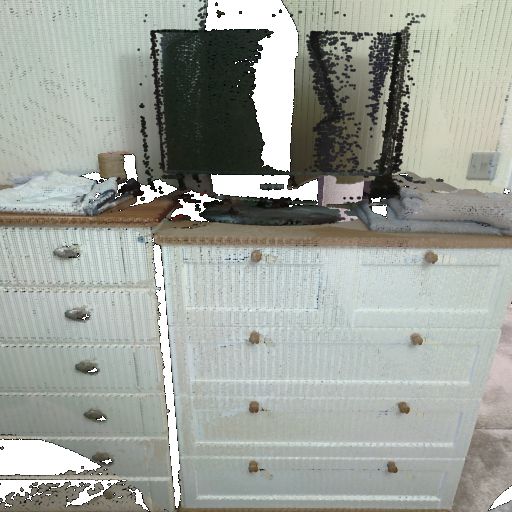}
        \caption*{w/o Verification}
    \end{subfigure}\hfill
    \begin{subfigure}[b]{0.32\linewidth}
        \includegraphics[width=\linewidth]{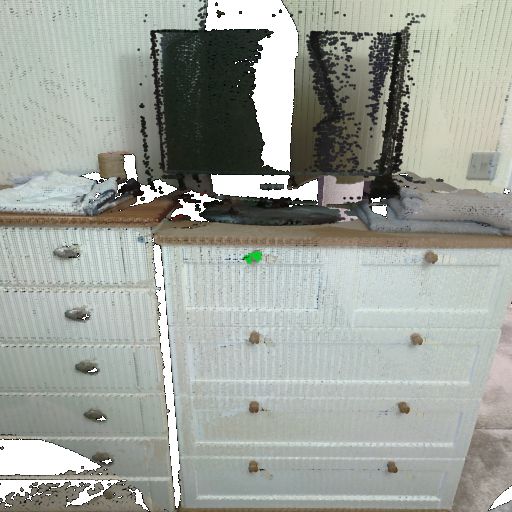}
        \caption*{Ours}
    \end{subfigure}\hfill
    \begin{subfigure}[b]{0.32\linewidth}
        \includegraphics[width=\linewidth]{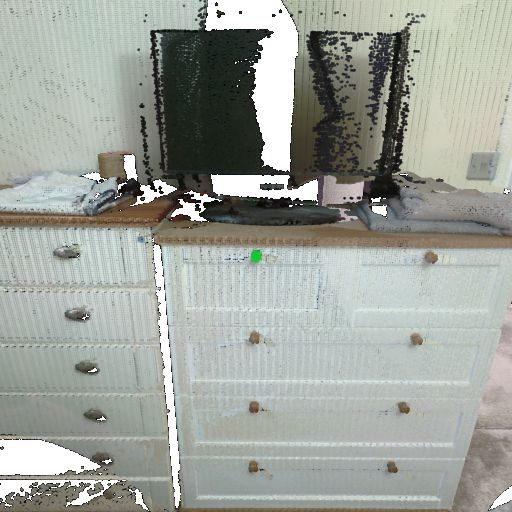}
        \caption*{GT}
    \end{subfigure}

    \vspace{1em}

    \begin{subfigure}[b]{0.32\linewidth}
        \includegraphics[width=\linewidth]{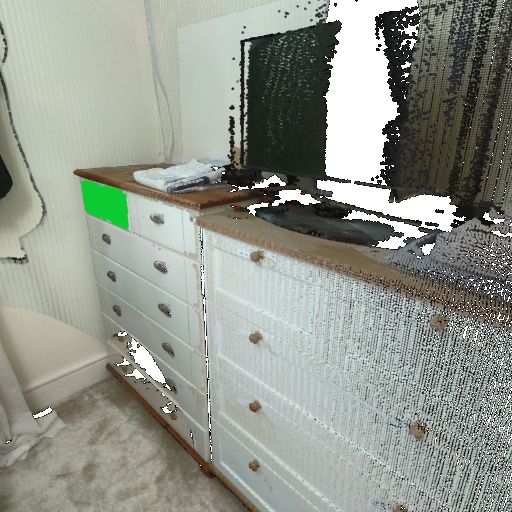}
        \caption*{Base}
    \end{subfigure}\hfill
    \begin{subfigure}[b]{0.32\linewidth}
        \includegraphics[width=\linewidth]{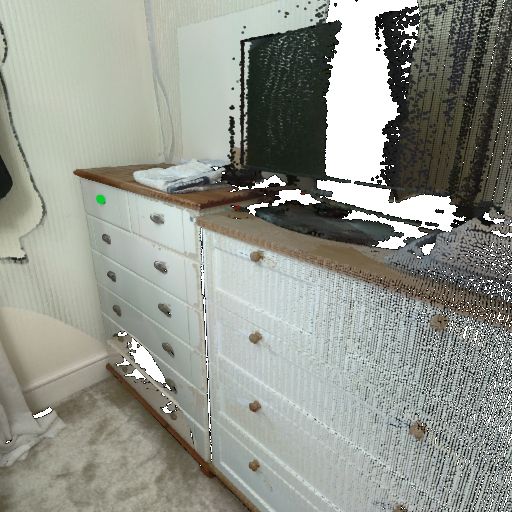}
        \caption*{w/o Multi-view}
    \end{subfigure}\hfill
    \begin{subfigure}[b]{0.32\linewidth}
        \includegraphics[width=\linewidth]{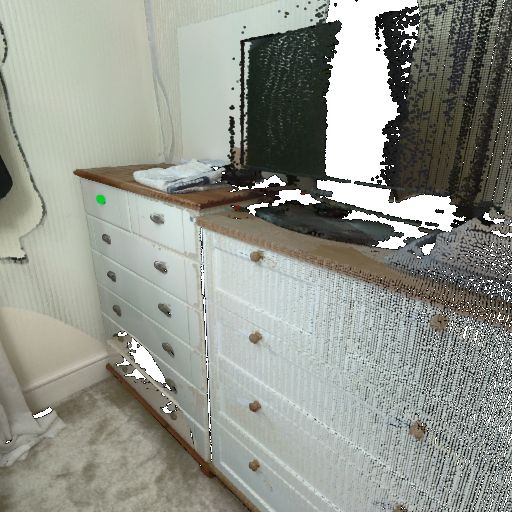}
        \caption*{w/o Temporal}
    \end{subfigure}

    \vspace{0.5em}

    \begin{subfigure}[b]{0.32\linewidth}
        \includegraphics[width=\linewidth]{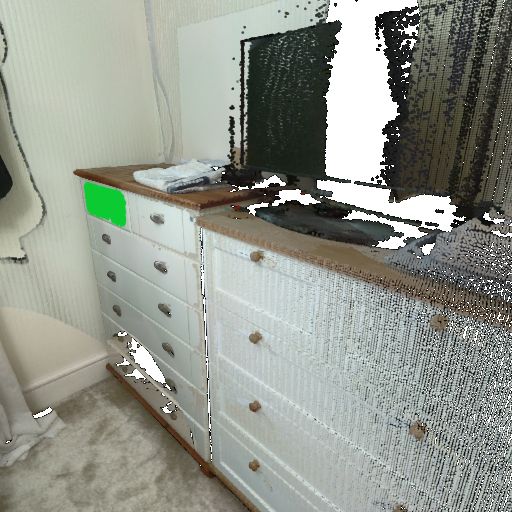}
        \caption*{w/o Verification}
    \end{subfigure}\hfill
    \begin{subfigure}[b]{0.32\linewidth}
        \includegraphics[width=\linewidth]{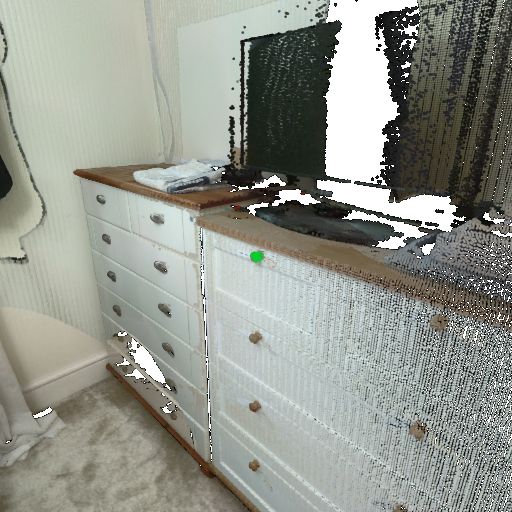}
        \caption*{Ours}
    \end{subfigure}\hfill
    \begin{subfigure}[b]{0.32\linewidth}
        \includegraphics[width=\linewidth]{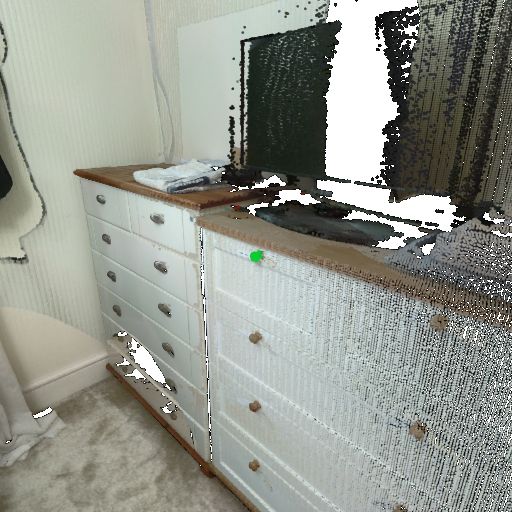}
        \caption*{GT}
    \end{subfigure}

    \caption{Visual comparisons for ablation study. From left to right, top to bottom: Base, w/o Multi-view, w/o Temporal, w/o Verification, Ours, and GT. Two different viewpoints are shown for the same test example. Input prompt: \texttt{Open the top left drawer of the cabinet with the TV on top}.}
    \label{fig:ablation}
\end{figure}

Fig.~\ref{fig:ablation} shows a further visual comparison for ablation models, which removes the temporal window, multi-sampling and verification, respectively. For input prompt: \texttt{Open the top left drawer of the cabinet with the TV on top}, the base model wrongly predicts expansive and inaccurate masks of the top left drawer of the left cabinet, but not the one with the TV on top, resulting in an mIoU of 0.000, as it fails to isolate the fine-grained functional part of the knob due to the limited size of selected frame candidates for correct spatial-temporal reasoning. Similarly, removing the temporal window or multi-sampling leads to mislocalization or missing detections entirely, though it can nearly locate a knob, but not the correct one.
However, after removing verification, the model will mistakenly treat the drawer as the target functional object instead, due to the absence of verification to check if the mask is the exact functional object type \texttt{knob}.
In contrast, our full UniFunc3D model achieves an mIoU of 0.523 by correctly identifying the precise target region, aligning closely with the ground truth. This visual evidence confirms that our unified reasoning-perception approach and temporal strategies are essential for robust 3D affordance grounding, particularly for fine-grained parts where simpler baselines often suffer from over-segmentation or localization failure.

\begin{figure}[t]
    \centering
    \begin{subfigure}[b]{0.32\linewidth}
        \includegraphics[width=\linewidth]{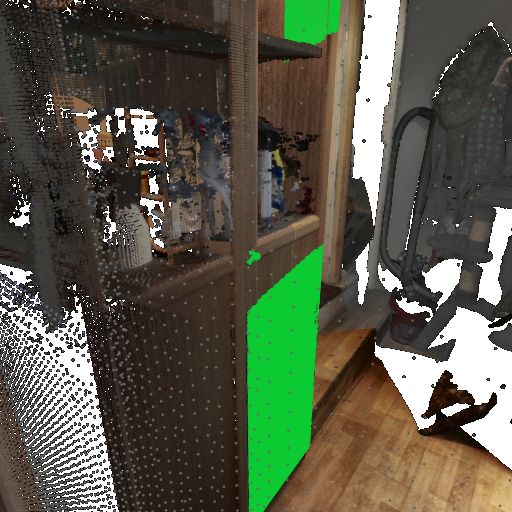}
        \caption*{Base}
    \end{subfigure}\hfill
    \begin{subfigure}[b]{0.32\linewidth}
        \includegraphics[width=\linewidth]{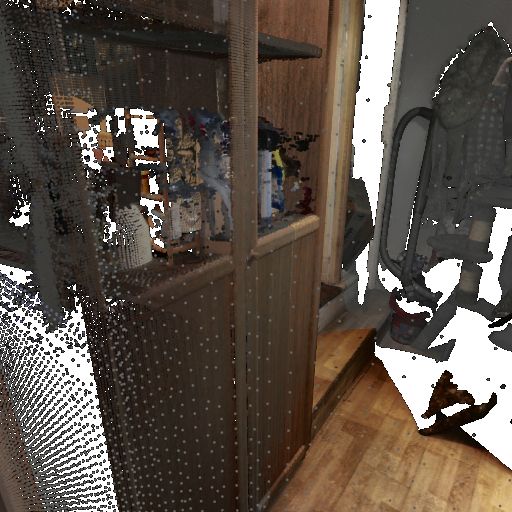}
        \caption*{w/o Multi-view}
    \end{subfigure}\hfill
    \begin{subfigure}[b]{0.32\linewidth}
        \includegraphics[width=\linewidth]{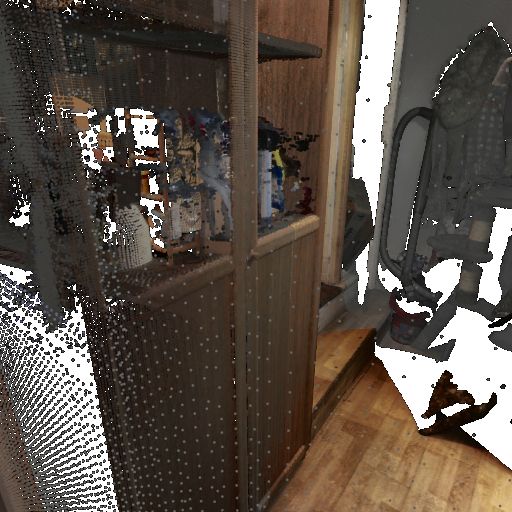}
        \caption*{w/o Temporal}
    \end{subfigure}

    \vspace{0.5em}

    \begin{subfigure}[b]{0.32\linewidth}
        \includegraphics[width=\linewidth]{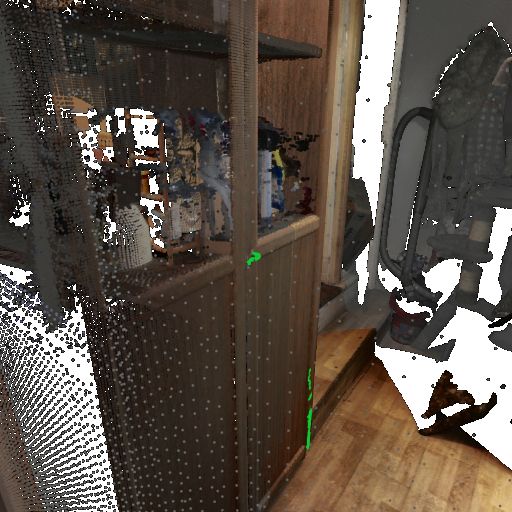}
        \caption*{w/o Verification}
    \end{subfigure}\hfill
    \begin{subfigure}[b]{0.32\linewidth}
        \includegraphics[width=\linewidth]{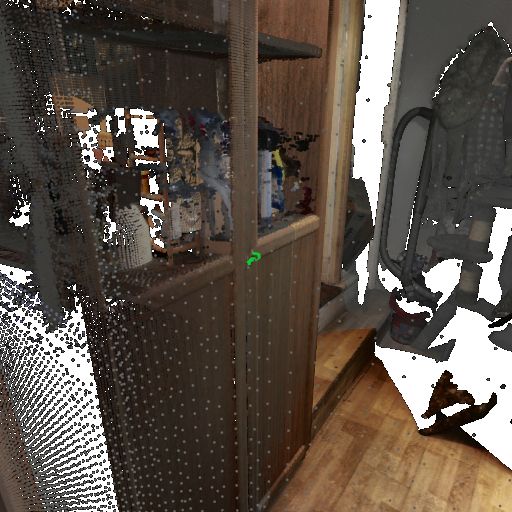}
        \caption*{Ours}
    \end{subfigure}\hfill
    \begin{subfigure}[b]{0.32\linewidth}
        \includegraphics[width=\linewidth]{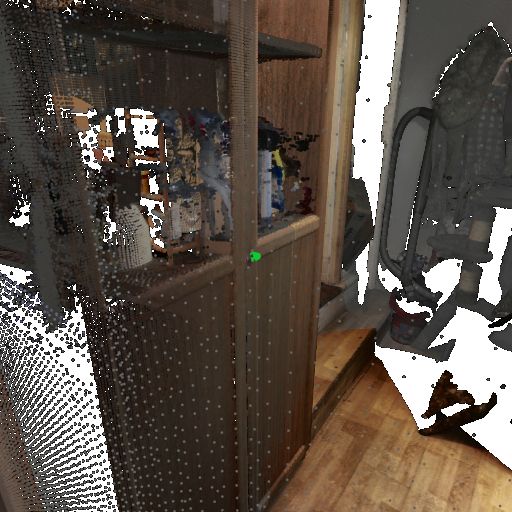}
        \caption*{GT}
    \end{subfigure}

    \vspace{1em}

    \begin{subfigure}[b]{0.32\linewidth}
        \includegraphics[width=\linewidth]{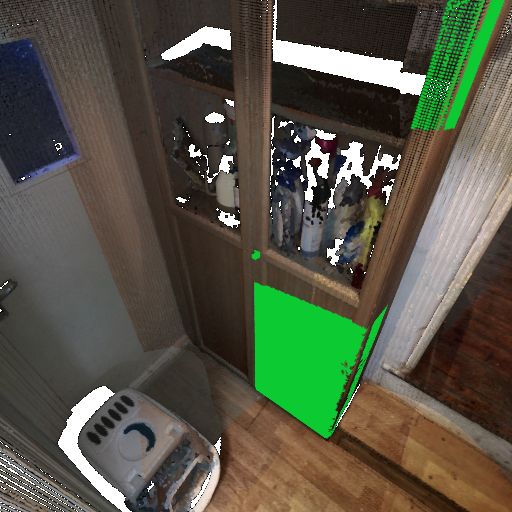}
        \caption*{Base}
    \end{subfigure}\hfill
    \begin{subfigure}[b]{0.32\linewidth}
        \includegraphics[width=\linewidth]{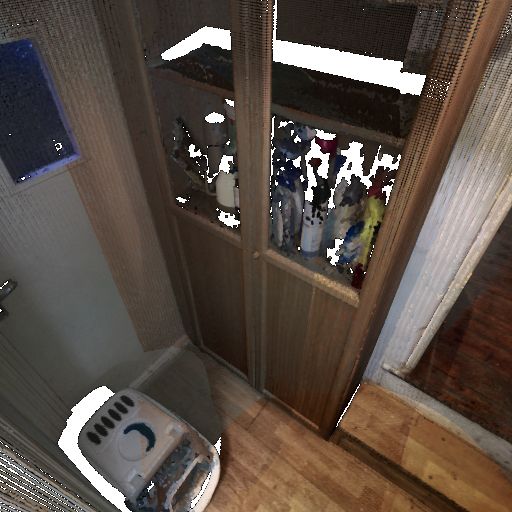}
        \caption*{w/o Multi-view}
    \end{subfigure}\hfill
    \begin{subfigure}[b]{0.32\linewidth}
        \includegraphics[width=\linewidth]{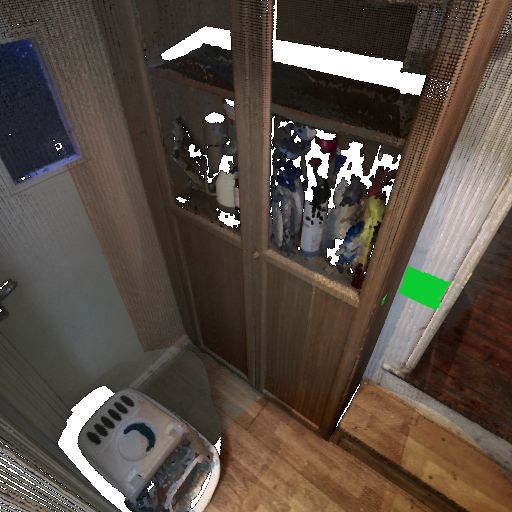}
        \caption*{w/o Temporal}
    \end{subfigure}

    \vspace{0.5em}

    \begin{subfigure}[b]{0.32\linewidth}
        \includegraphics[width=\linewidth]{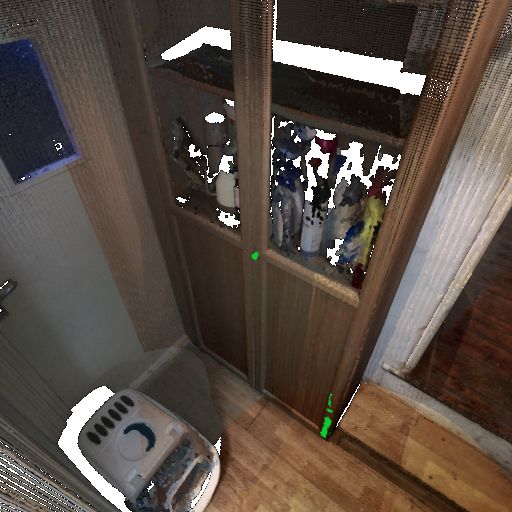}
        \caption*{w/o Verification}
    \end{subfigure}\hfill
    \begin{subfigure}[b]{0.32\linewidth}
        \includegraphics[width=\linewidth]{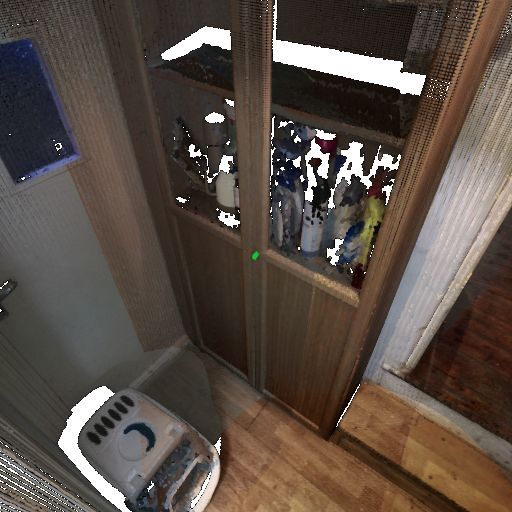}
        \caption*{Ours}
    \end{subfigure}\hfill
    \begin{subfigure}[b]{0.32\linewidth}
        \includegraphics[width=\linewidth]{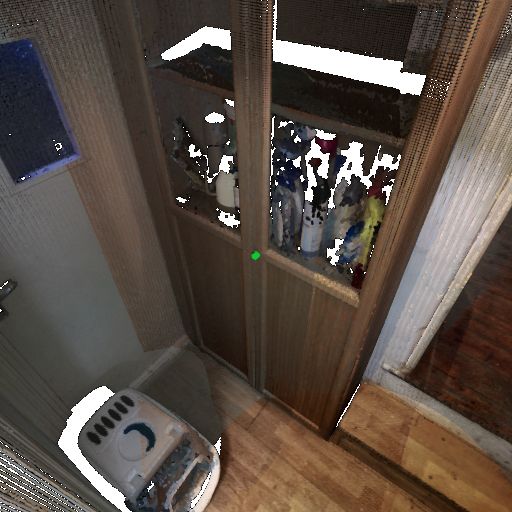}
        \caption*{GT}
    \end{subfigure}

    \caption{Visual comparisons for ablation study. From left to right, top to bottom: Base, w/o Multi-view, w/o Temporal, w/o Verification, Ours, and GT. Two different viewpoints are shown for the same test example. Input prompt: \texttt{Open the right door of the wood and glass display cabinet next to the pet transport box}.}
    \label{fig:ablation2}
\end{figure}

Similarly, Fig.~\ref{fig:ablation2} shows another further visual comparison for ablation models, given an input text \texttt{Open the right door of the wood and glass display cabinet next to the pet transport box}. We can see that without multi-sampling and temporal window the model cannot locate the correct \texttt{knob} (both mIoU: 0.00) due to the limited number of selected frame candidates with potential targets and lack good spatial-temporal reasoning during this process. While the ablation model without verification can find the correct knob, it will treat some other points as the target object (at the bottom corner of the cabinet), resulting in an mIoU of 0.058. Our method can achieve the best results (mIoU: 0.476) with the help of sufficient spatial-temporal reasoning and the final verification. These examples further demonstrate the effectiveness of our proposed pipelines.

\section{Additional Related Work}
\noindent \textbf{Open-vocabulary 3D segmentation.}
Open-vocabulary 3D (OV-3D) segmentation methods~\cite{peng2023openscene,takmaz2024openmask3d,yin2024sai3d,ton2024zeroshotdualpath,nguyen2024open3dis,huang2024openins3d} aim to segment objects in 3D scenes using natural language descriptions.
These approaches typically combine 3D proposal modules for predicting masks from point clouds~\cite{choy2019minkowski,schult2023mask3d} with 2D modules that extract masks from multi-view RGB images using VLMs~\cite{clip,liu2023visual} and segmentation models~\cite{kirillov2023segany,zou2023segment}.
Segmentation is achieved through 2D-3D fusion based on mask agreement~\cite{takmaz2024openmask3d,tai2024opensam3d,ton2024zeroshotdualpath} or learnable pooling~\cite{nguyen2024open3dis,peng2023openscene}.
Language-guided radiance field methods~\cite{kerr2023lerf,qin2024langsplat,engelmann2024opennerf} can also perform OV3DS but require scene-specific training.
Empirical results on SceneFun3D~\cite{delitzas2024scenefun3d} demonstrate that these methods struggle with affordance segmentation, as they rely on modules pre-trained on 3D datasets~\cite{chang2017matterport3d,dai2017scannet} biased toward large furniture rather than small functional parts, and they interpret concise descriptions with explicit object names rather than implicit, context-dependent task descriptions.
Moreover, these methods lack mechanisms to disambiguate multiple instances of the same object class (\textit{e.g.}, selecting the correct cabinet when multiple cabinets exist) or to leverage multi-frame temporal context for spatial referring expressions.
UniFunc3D addresses these limitations by using an MLLM to interpret complex task descriptions with visual evidence and performing video-based temporal grounding to resolve spatial ambiguities across frames. 

\jynew{
\textbf{Applications on 3D functionality scene graph.} FunGraph~\cite{fungraph} augments 3D scene graphs with functional elements for language-prompted interaction.
OpenFunGraph~\cite{openfungraph} predicts open-vocabulary functional 3D scene graphs and introduces FunGraph3D, and KeySG~\cite{keysg} uses keyframes and hierarchical retrieval to scale scene-graph reasoning to larger environments. These methods emphasize persistent scene memory and object-functional-element relationships, while our target is query-conditioned 3D mask prediction on SceneFun3D and can not directly be address by them. 
}

\section{Additional Qualitative Results}
\label{sec:addition_results}

\begin{figure}[h]
\centering
\setlength{\tabcolsep}{1pt}
\renewcommand{\arraystretch}{0.5}
\begin{tabular}{c@{\hspace{3pt}}ccccc}

  &
  \parbox[c]{\imgw}{\centering\scriptsize\texttt{Unplug the table lamp on the nightstand}} &
  \parbox[c]{\imgw}{\centering\scriptsize\texttt{Open the top drawer of the nightstand next to the closet}} &
  \parbox[c]{\imgw}{\centering\scriptsize\texttt{Open the left window part}} &
  \parbox[c]{\imgw}{\centering\scriptsize\texttt{Open the right door of the wooden closet}} &
  \parbox[c]{\imgw}{\centering\scriptsize\texttt{Turn on the ceiling light}} \\[4pt]
    \rlabel{AffordBot} &
    \includegraphics[width=\imgw]{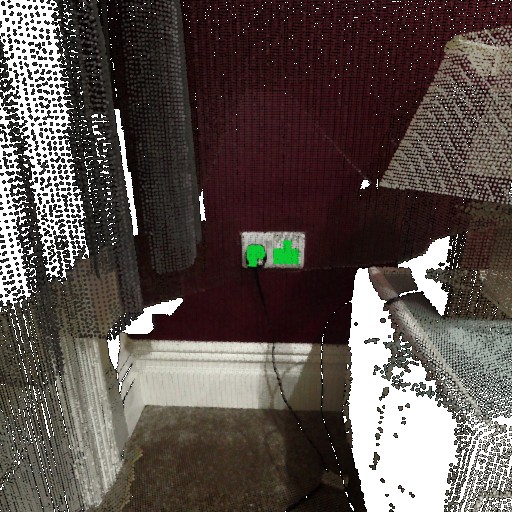} &
    \includegraphics[width=\imgw]{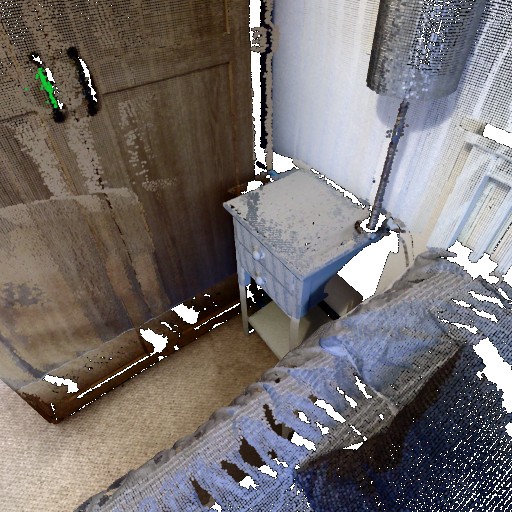} &
    \includegraphics[width=\imgw]{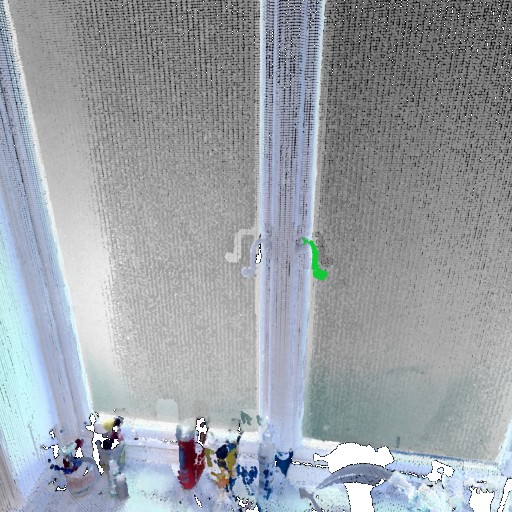} &
    \includegraphics[width=\imgw]{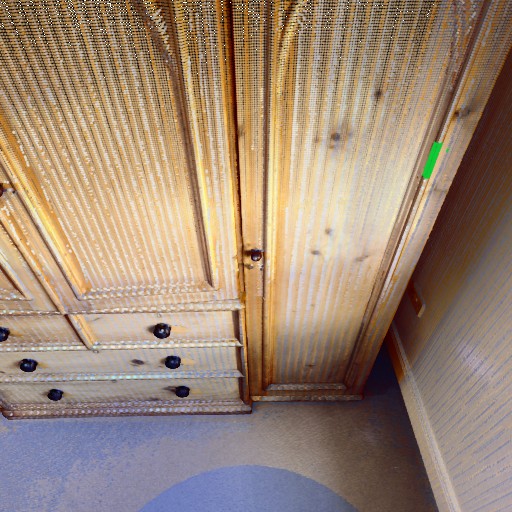} &
    \includegraphics[width=\imgw]{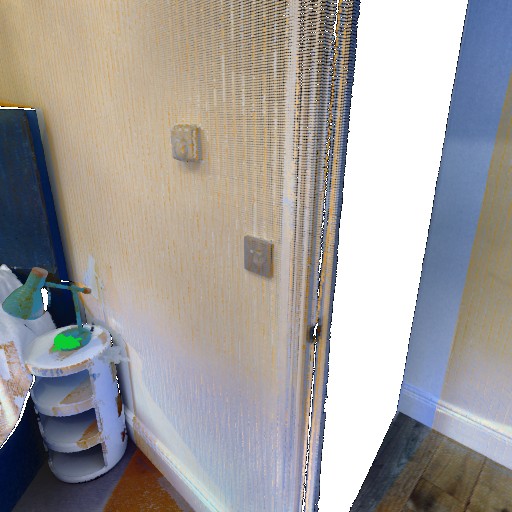} \\[2pt]
  \rlabel{Fun3DU} &
    \includegraphics[width=\imgw]{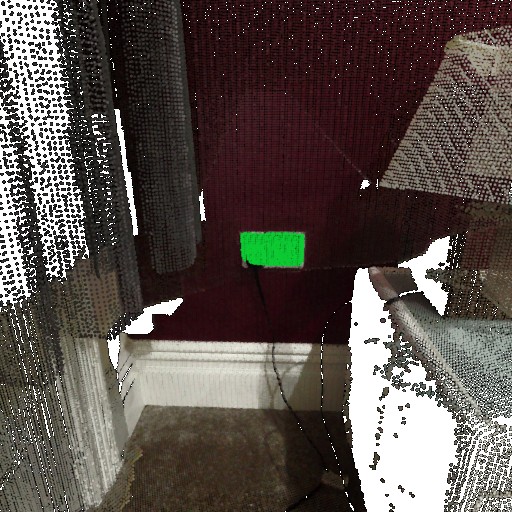} &
    \includegraphics[width=\imgw]{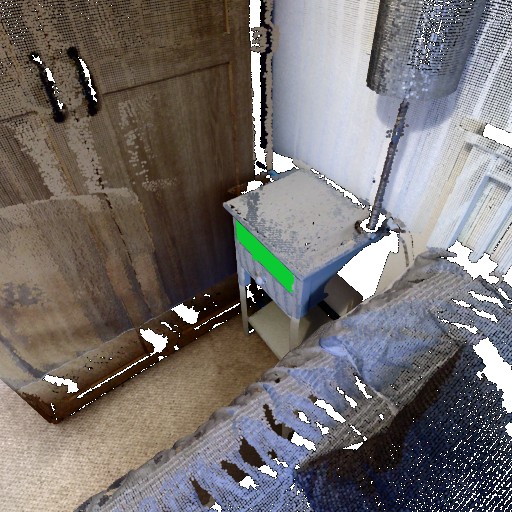} &
    \includegraphics[width=\imgw]{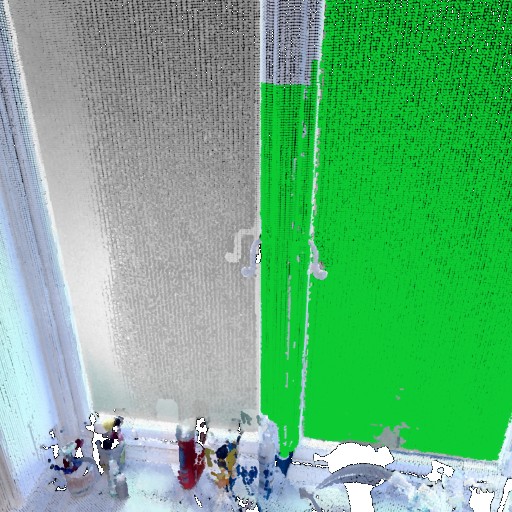} &
    \includegraphics[width=\imgw]{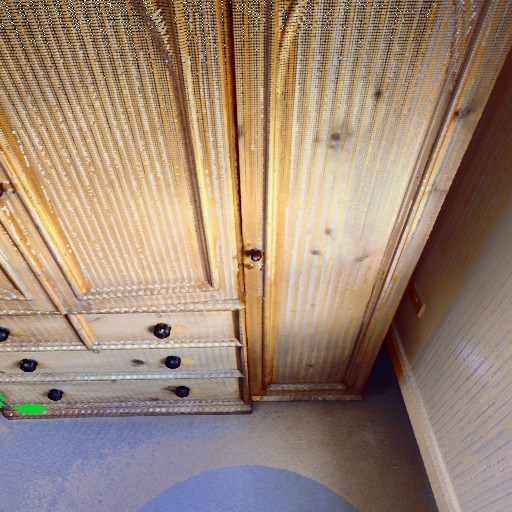} &
    \includegraphics[width=\imgw]{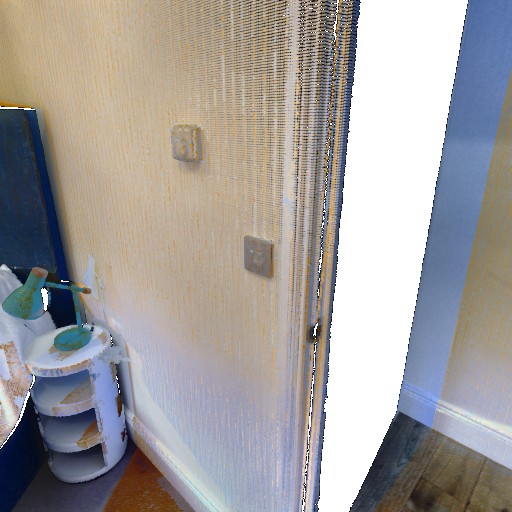} \\[2pt]
  \rlabel{Ours} &
    \includegraphics[width=\imgw]{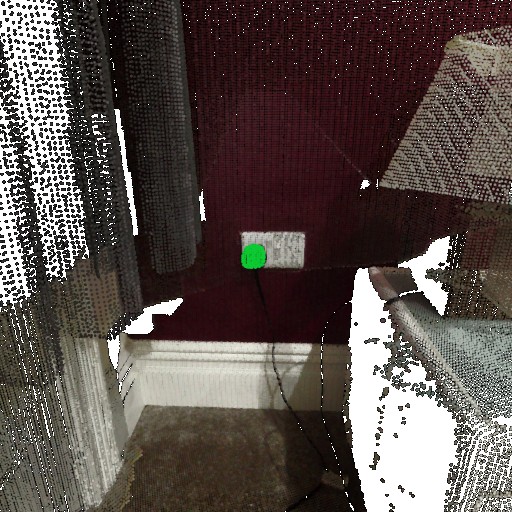} &
    \includegraphics[width=\imgw]{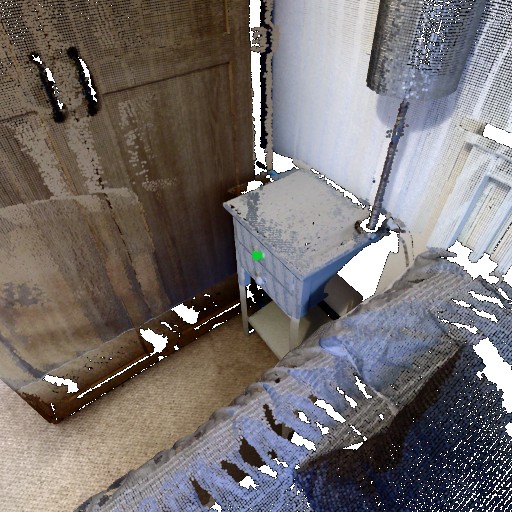} &
    \includegraphics[width=\imgw]{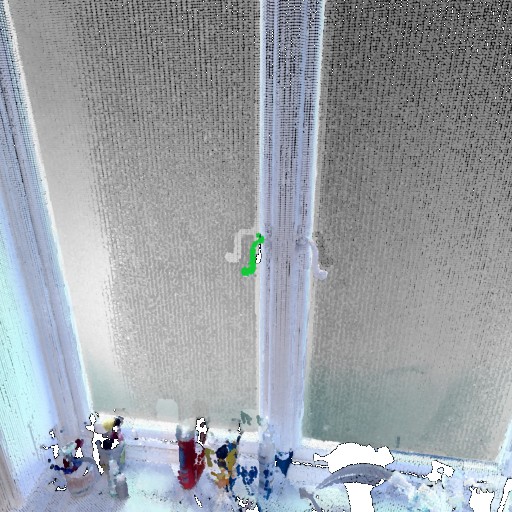} &
    \includegraphics[width=\imgw]{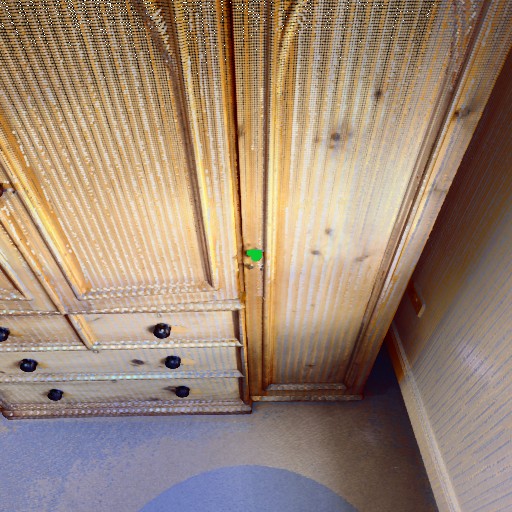} &
    \includegraphics[width=\imgw]{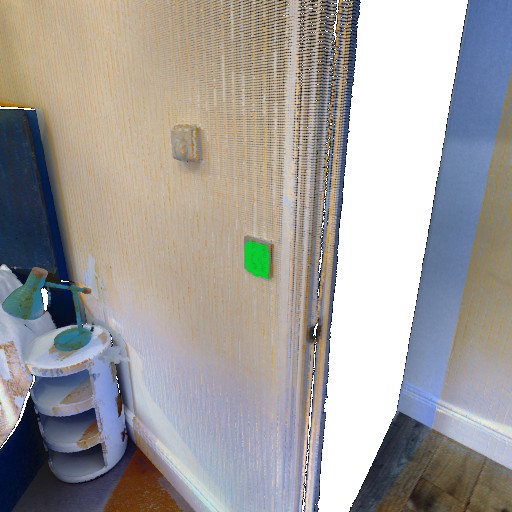} \\[2pt]
  \rlabel{GT} &
    \includegraphics[width=\imgw]{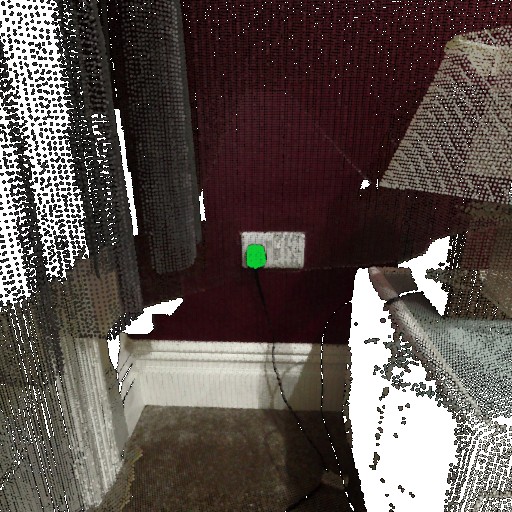} &
    \includegraphics[width=\imgw]{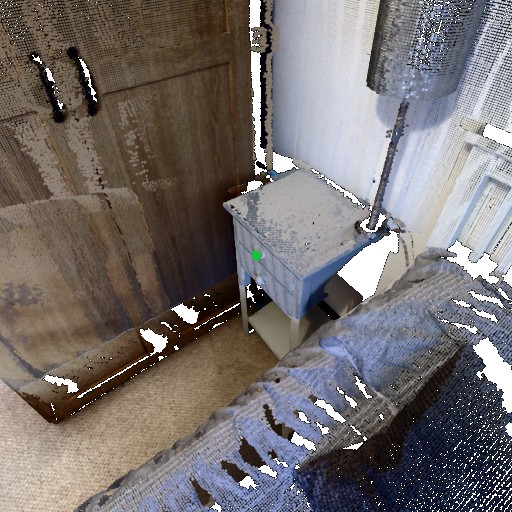} &
    \includegraphics[width=\imgw]{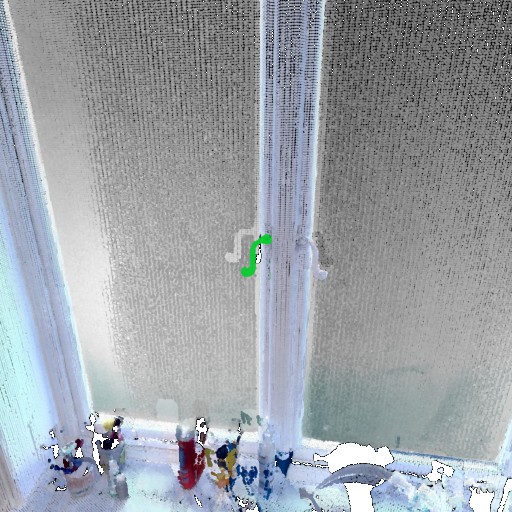} &
    \includegraphics[width=\imgw]{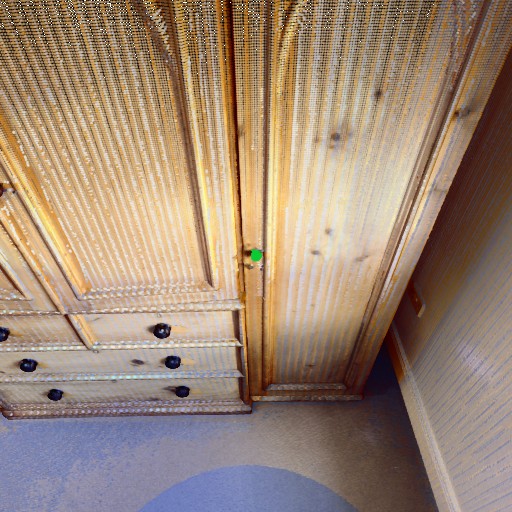} &
    \includegraphics[width=\imgw]{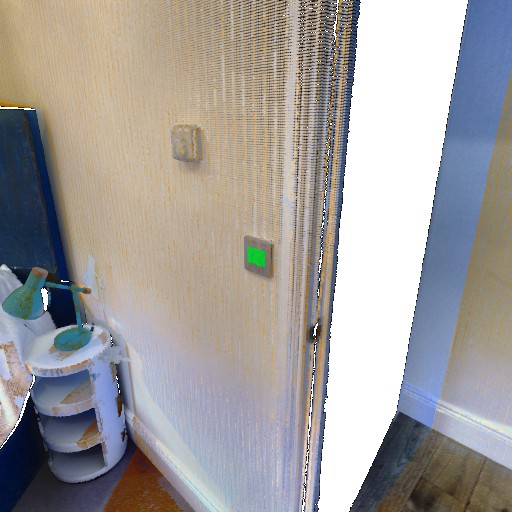} \\[2pt]
\end{tabular}
\vspace{-3mm}
\caption{\textbf{Qualitative comparison.} We show results for five representative queries (columns) across four methods (rows). GT denotes ground truth.}
\label{fig:supp1}
\end{figure}

\begin{figure}[t]
\centering
\setlength{\tabcolsep}{1pt}
\renewcommand{\arraystretch}{0.5}
\begin{tabular}{c@{\hspace{3pt}}ccccc}

  &
  \parbox[c]{\imgw}{\centering\scriptsize\texttt{Open the bottom drawer of the nightstand with the books on top}} &
  \parbox[c]{\imgw}{\centering\scriptsize\texttt{Open the right middle drawer of the dressing table}} &
  \parbox[c]{\imgw}{\centering\scriptsize\texttt{Open the middle left drawer of the dressing table}} &
  \parbox[c]{\imgw}{\centering\scriptsize\texttt{Open the bottom drawer of the nightstand}} &
  \parbox[c]{\imgw}{\centering\scriptsize\texttt{Open the window to the left of the vanity table}} \\[4pt]
    \rlabel{AffordBot} &
    \includegraphics[width=\imgw]{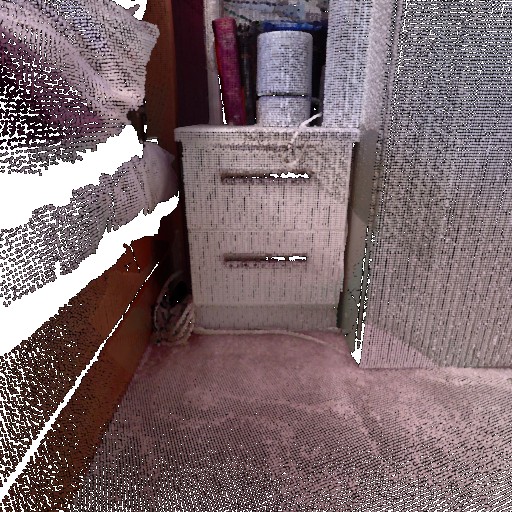} &
    \includegraphics[width=\imgw]{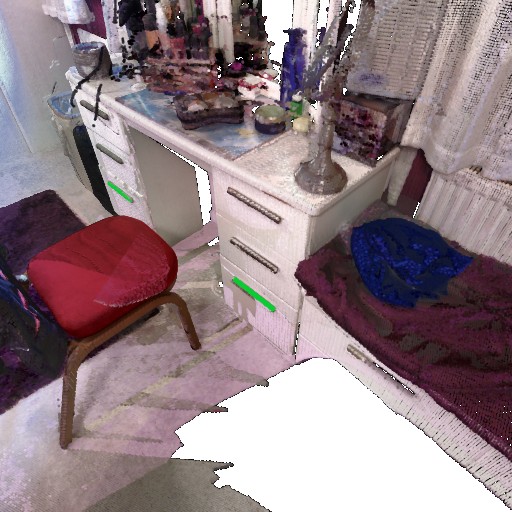} &
    \includegraphics[width=\imgw]{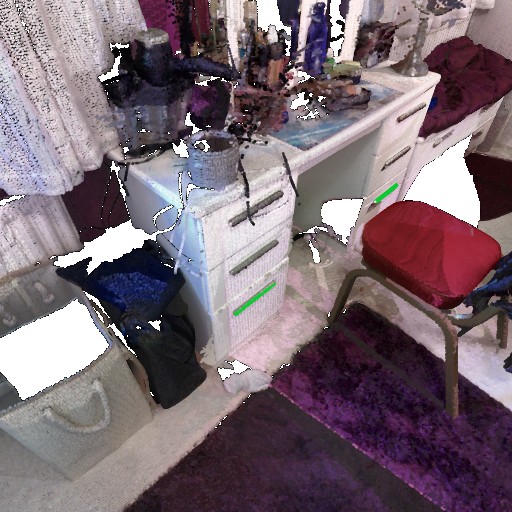} &
    \includegraphics[width=\imgw]{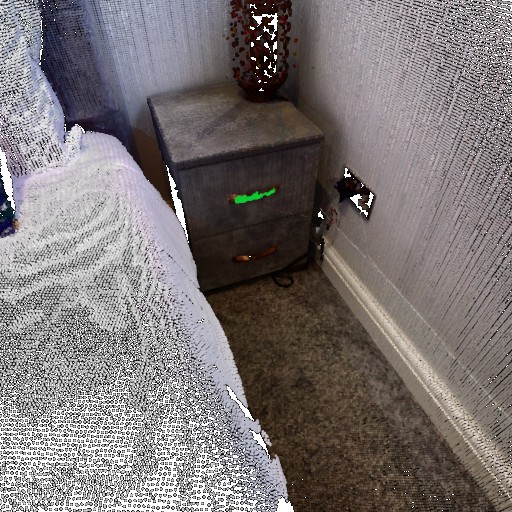} &
    \includegraphics[width=\imgw]{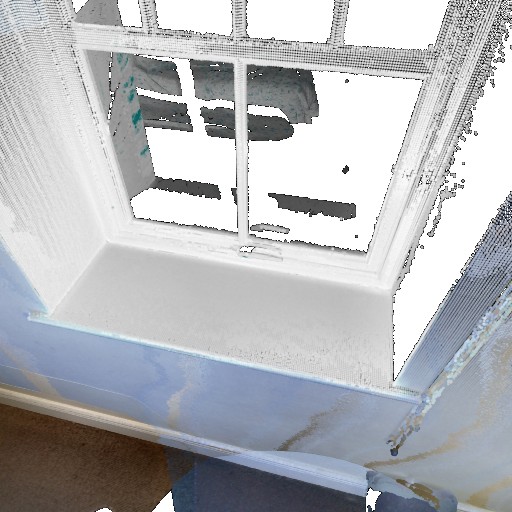} \\[2pt]
  \rlabel{Fun3DU} &
    \includegraphics[width=\imgw]{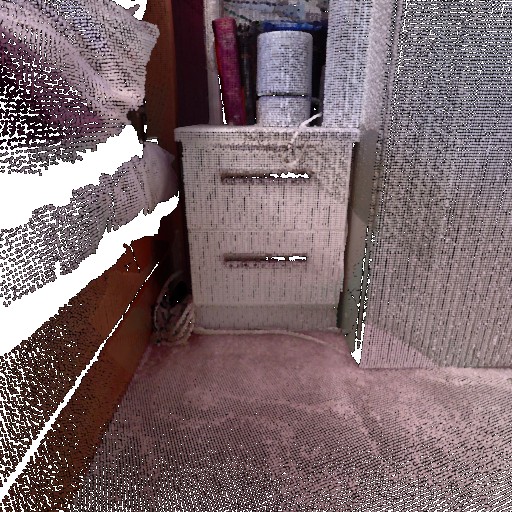} &
    \includegraphics[width=\imgw]{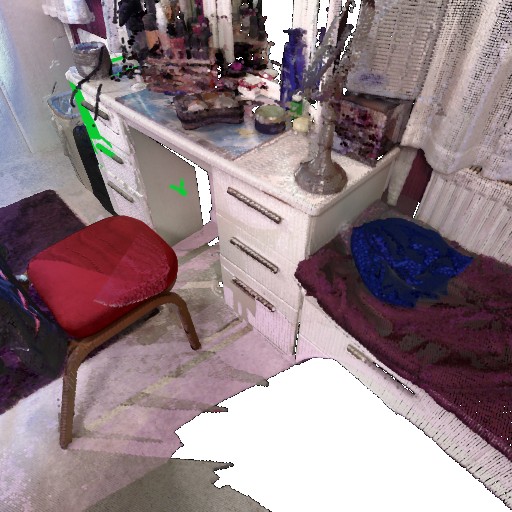} &
    \includegraphics[width=\imgw]{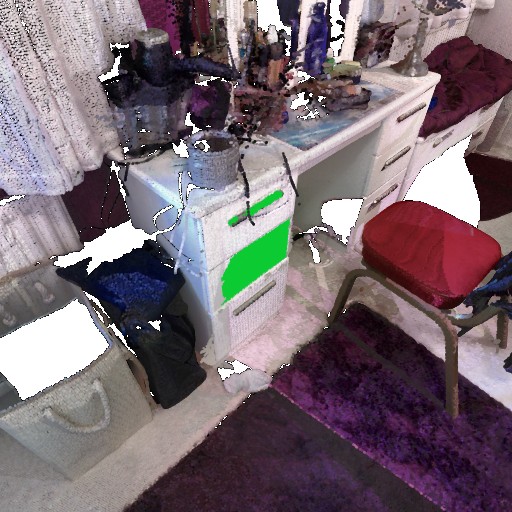} &
    \includegraphics[width=\imgw]{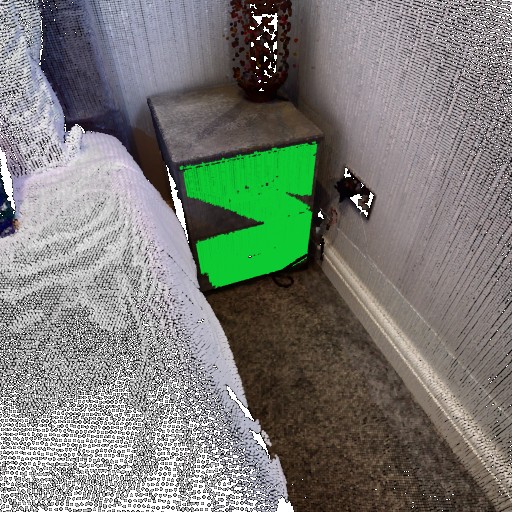} &
    \includegraphics[width=\imgw]{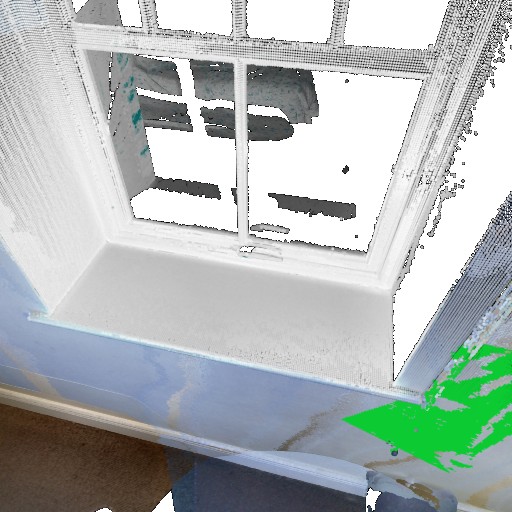} \\[2pt]
  \rlabel{Ours} &
    \includegraphics[width=\imgw]{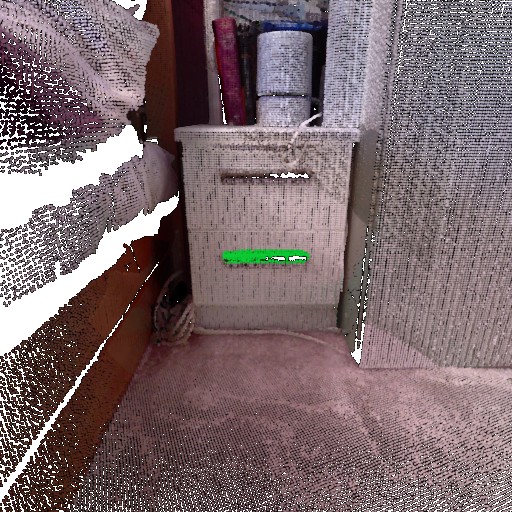} &
    \includegraphics[width=\imgw]{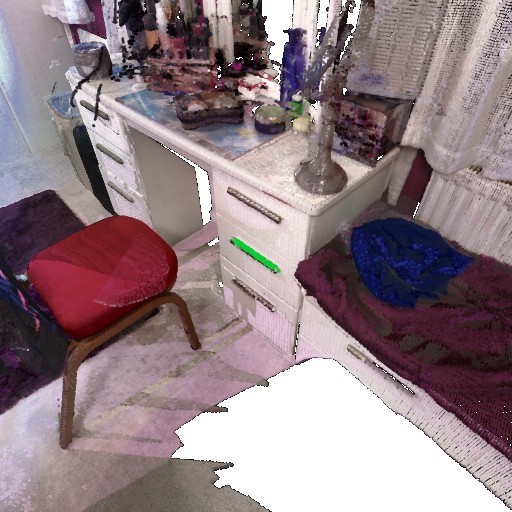} &
    \includegraphics[width=\imgw]{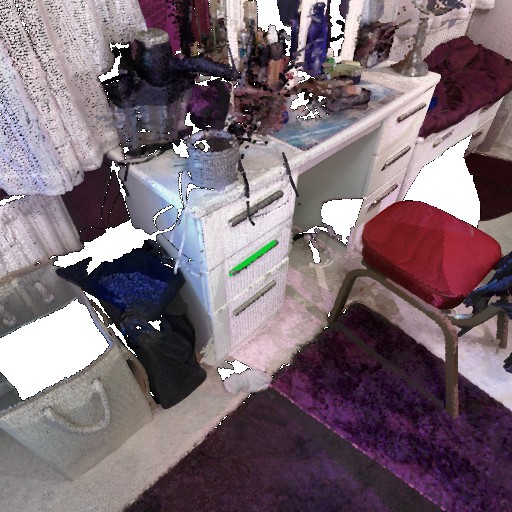} &
    \includegraphics[width=\imgw]{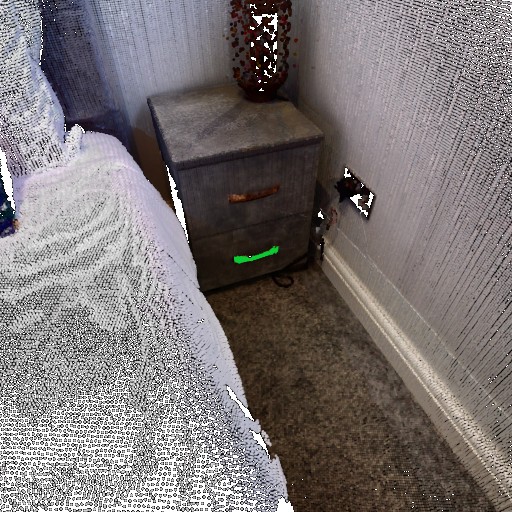} &
    \includegraphics[width=\imgw]{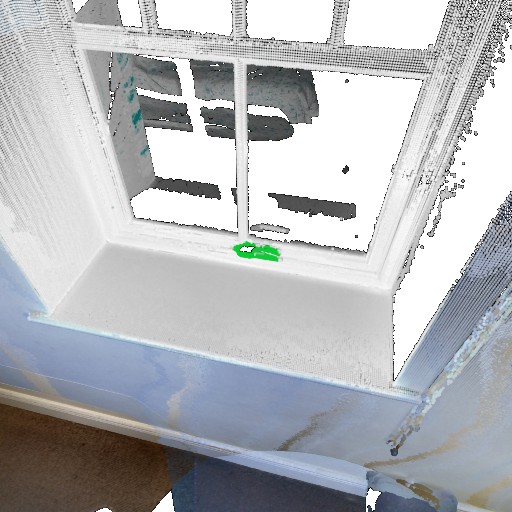} \\[2pt]
  \rlabel{GT} &
    \includegraphics[width=\imgw]{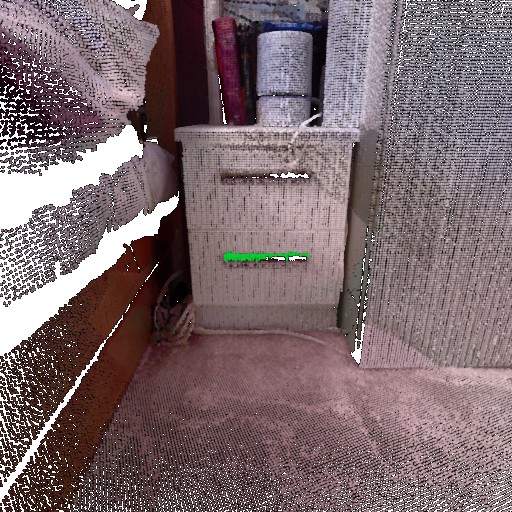} &
    \includegraphics[width=\imgw]{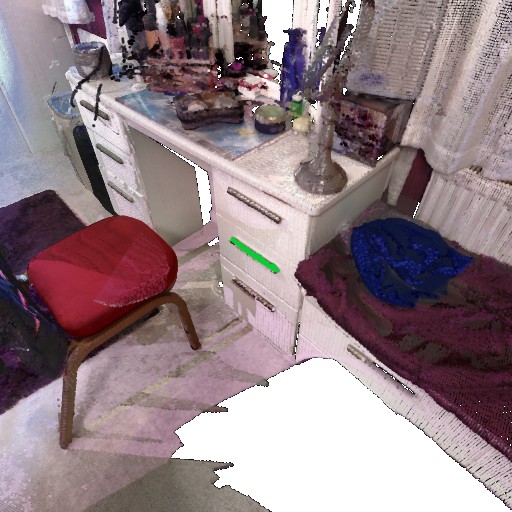} &
    \includegraphics[width=\imgw]{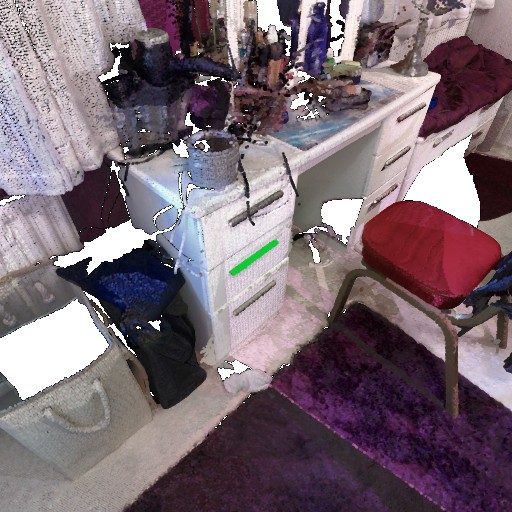} &
    \includegraphics[width=\imgw]{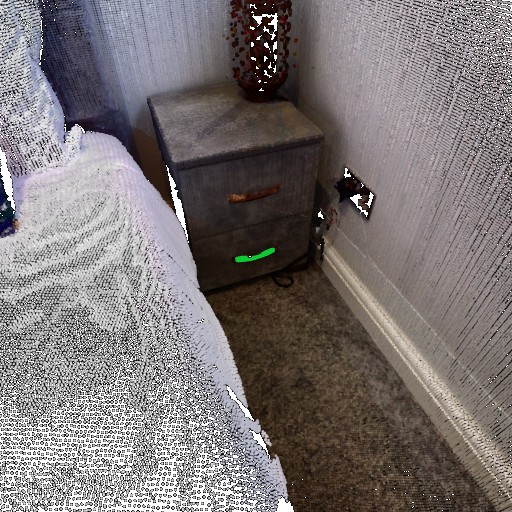} &
    \includegraphics[width=\imgw]{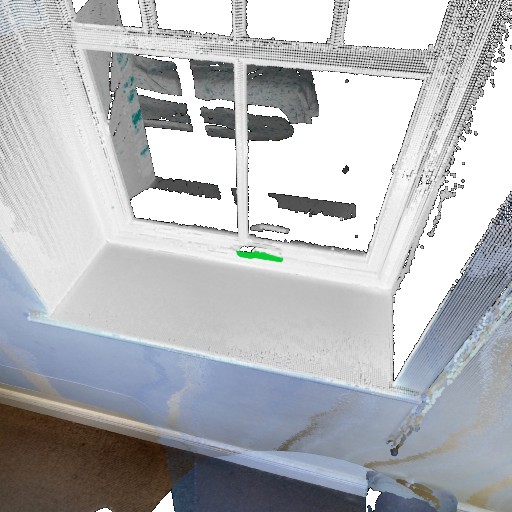} \\[2pt]
\end{tabular}
\vspace{-3mm}
\caption{\textbf{Qualitative comparison.} We show results for five representative queries (columns) across four methods (rows). GT denotes ground truth.}
\label{fig:supp2}
\end{figure}

\begin{figure}[t]
\centering
\setlength{\tabcolsep}{1pt}
\renewcommand{\arraystretch}{0.5}
\begin{tabular}{c@{\hspace{3pt}}ccccc}

  &
  \parbox[c]{\imgw}{\centering\scriptsize\texttt{Open the top left drawer of the closet next to the door}} &
  \parbox[c]{\imgw}{\centering\scriptsize\texttt{Open the right door of the wooden closet}} &
  \parbox[c]{\imgw}{\centering\scriptsize\texttt{Turn on the ceiling light}} &
  \parbox[c]{\imgw}{\centering\scriptsize\texttt{Open the right cabinet door with the TV on top}} &
  \parbox[c]{\imgw}{\centering\scriptsize\texttt{Open the bottom drawer of the wooden nightstand with the books on top}} \\[4pt]
    \rlabel{AffordBot} &
    \includegraphics[width=\imgw]{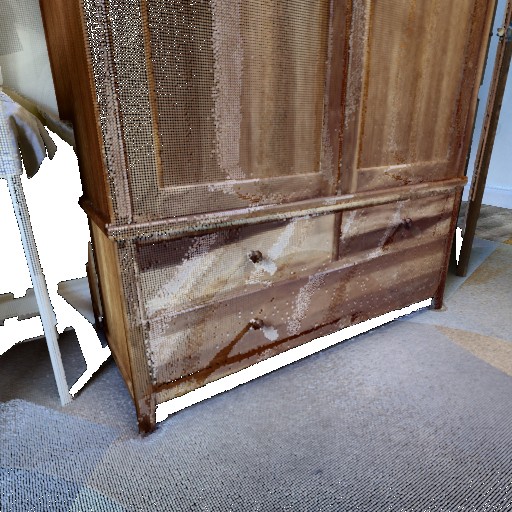} &
    \includegraphics[width=\imgw]{selected_results/affordbot/434897_009cc1a0-efc0-44a1-8808-6429e11fc5c1/view_8.jpg} &
    \includegraphics[width=\imgw]{selected_results/affordbot/434897_c6516dc2-acfd-49fc-bc46-6804d604ea63/view_3.jpg} &
    \includegraphics[width=\imgw]{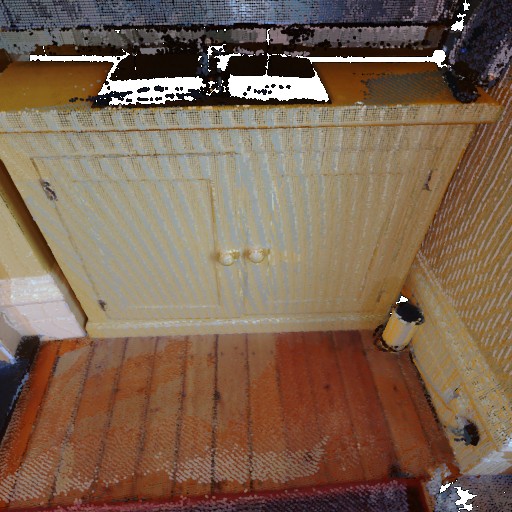} &
    \includegraphics[width=\imgw]{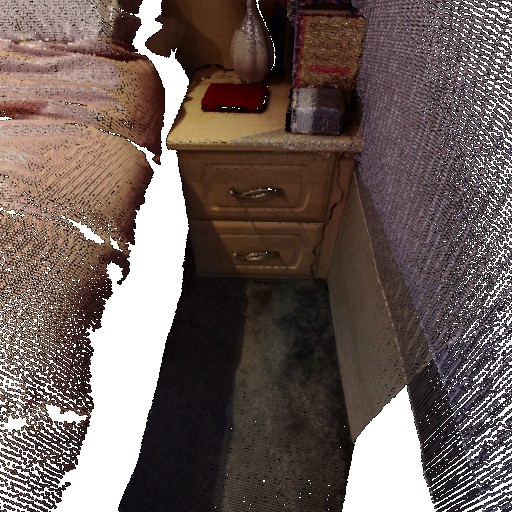} \\[2pt]
  \rlabel{Fun3DU} &
    \includegraphics[width=\imgw]{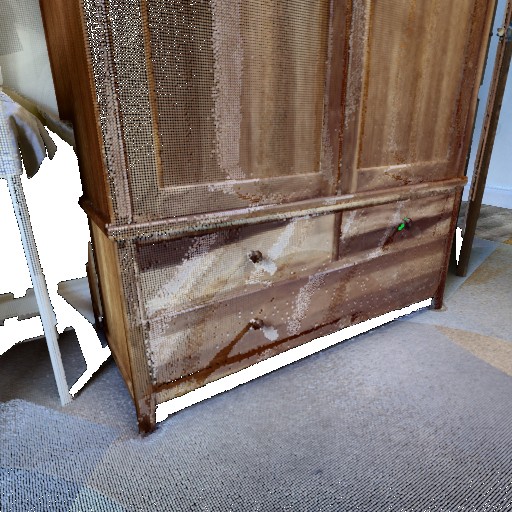} &
    \includegraphics[width=\imgw]{selected_results/fun3du/434897_009cc1a0-efc0-44a1-8808-6429e11fc5c1/view_8.jpg} &
    \includegraphics[width=\imgw]{selected_results/fun3du/434897_c6516dc2-acfd-49fc-bc46-6804d604ea63/view_3.jpg} &
    \includegraphics[width=\imgw]{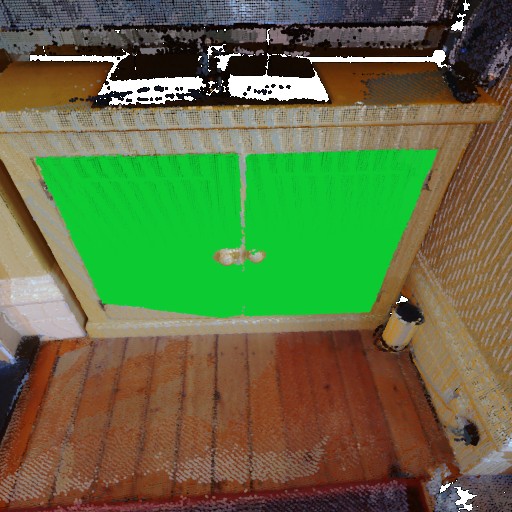} &
    \includegraphics[width=\imgw]{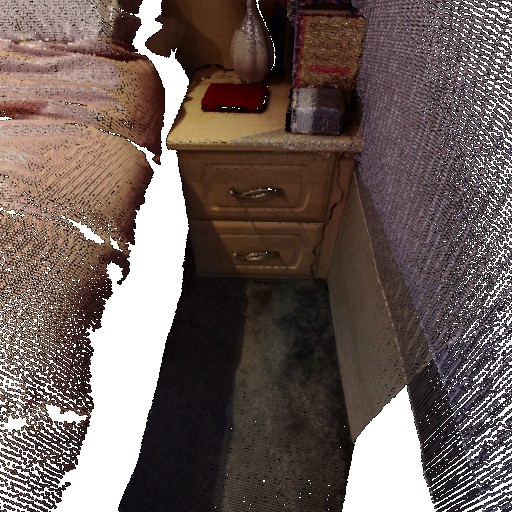} \\[2pt]
  \rlabel{Ours} &
    \includegraphics[width=\imgw]{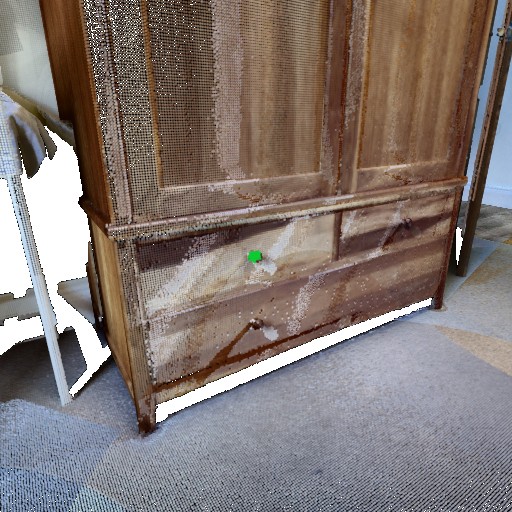} &
    \includegraphics[width=\imgw]{selected_results/ours/434897_009cc1a0-efc0-44a1-8808-6429e11fc5c1/view_8.jpg} &
    \includegraphics[width=\imgw]{selected_results/ours/434897_c6516dc2-acfd-49fc-bc46-6804d604ea63/view_3.jpg} &
    \includegraphics[width=\imgw]{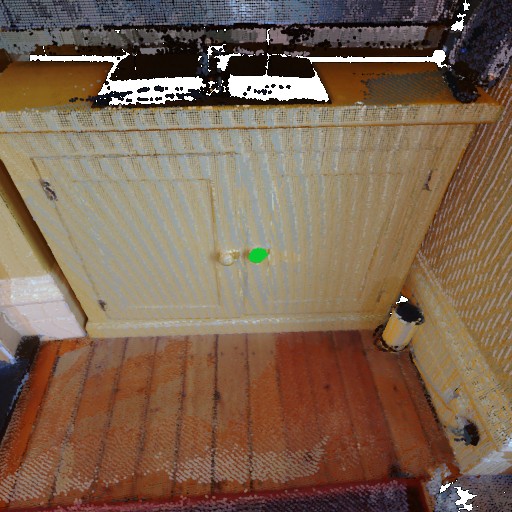} &
    \includegraphics[width=\imgw]{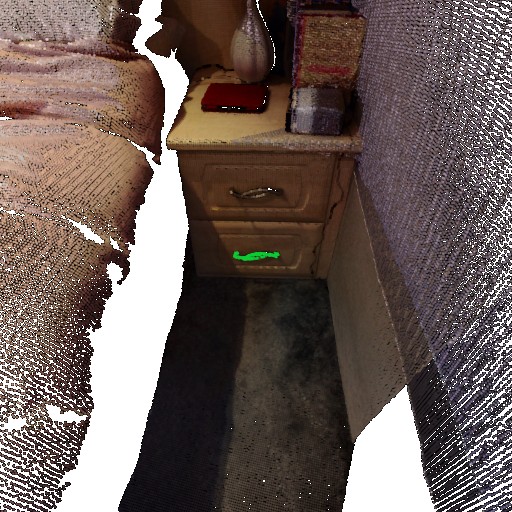} \\[2pt]
  \rlabel{GT} &
    \includegraphics[width=\imgw]{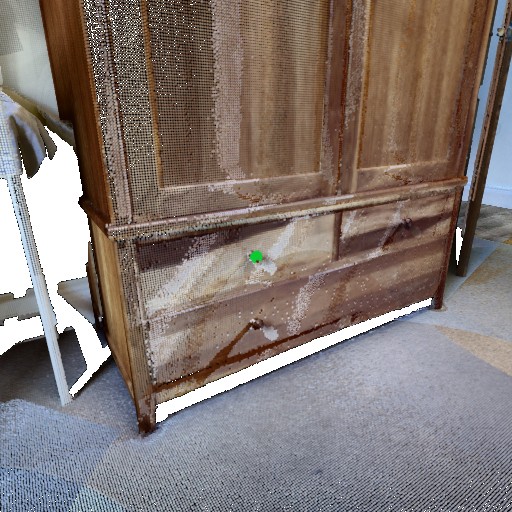} &
    \includegraphics[width=\imgw]{selected_results/gt/434897_009cc1a0-efc0-44a1-8808-6429e11fc5c1/view_8.jpg} &
    \includegraphics[width=\imgw]{selected_results/gt/434897_c6516dc2-acfd-49fc-bc46-6804d604ea63/view_3.jpg} &
    \includegraphics[width=\imgw]{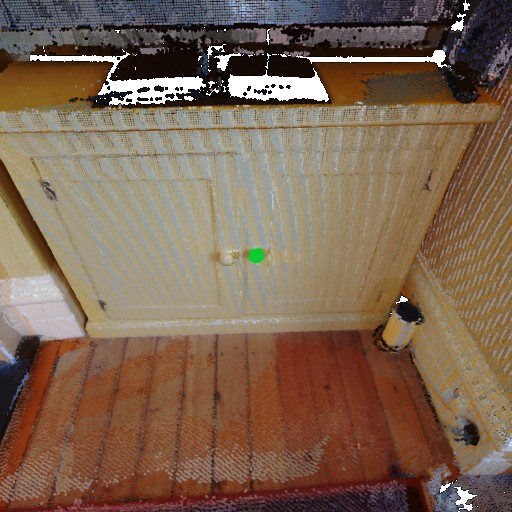} &
    \includegraphics[width=\imgw]{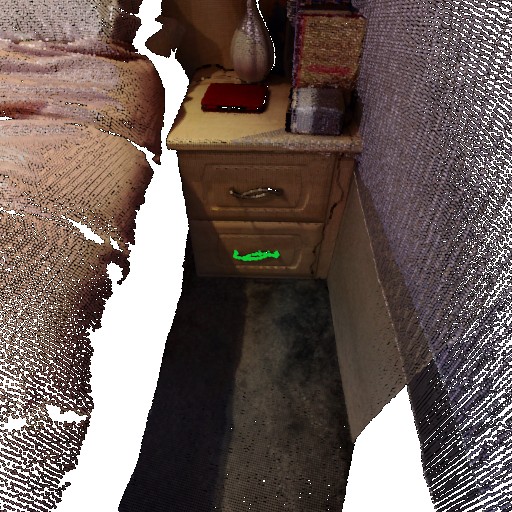} \\[2pt]
\end{tabular}
\vspace{-3mm}
\caption{\textbf{Qualitative comparison.} We show results for five representative queries (columns) across four methods (rows). GT denotes ground truth.}
\label{fig:supp3}
\end{figure}

Fig.~\ref{fig:supp1}, Fig.~\ref{fig:supp2} and Fig.~\ref{fig:supp3} show additional qualitative results. Our method outperforms other baselines in various scenes with different functional objects. Specifically, as illustrated in the visual comparisons, UniFunc3D exhibits superior localization accuracy for fine-grained parts, such as cabinet handles and specific drawer sections, where multi-stage baselines like Fun3DU often suffer from over-segmentation. Furthermore, in scenarios involving complex spatial relationships, such as distinguishing between multiple adjacent drawers, our unified reasoning-perception approach accurately identifies the target functional region, whereas training-based methods like AffordBot occasionally fail to generate precise masks for small-scale objects. These results qualitatively validate that our unified architecture not only improves efficiency but also provides more robust and semantically consistent 3D affordance grounding across diverse indoor environments.

\clearpage

\clearpage

\clearpage

\clearpage
\section*{NeurIPS Paper Checklist}

\begin{enumerate}

\item {\bf Claims}
    \item[] Question: Do the main claims made in the abstract and introduction accurately reflect the paper's contributions and scope?
    \item[] Answer: \answerYes{}
    \item[] Justification: The abstract claims state-of-the-art performance with a 59.9\% relative mIoU improvement on SceneFun3D, and this is directly substantiated by the quantitative results in Tab.~1 (Sec.~4.2) comparing against all prior methods across both evaluation splits.

\item {\bf Limitations}
    \item[] Question: Does the paper discuss the limitations of the work performed by the authors?
    \item[] Answer: \answerYes{}
    \item[] Justification: A dedicated ``Limitations and Future Work'' section (Appendix Sec.~4) discusses challenges with extremely small functional parts ($<$0.1\% of image area) and severe occlusion, and identifies directions for future work.

\item {\bf Theory assumptions and proofs}
    \item[] Question: For each theoretical result, does the paper provide the full set of assumptions and a complete (and correct) proof?
    \item[] Answer: \answerNA{}
    \item[] Justification: The paper presents an empirical systems contribution with no theoretical results, theorems, or formal proofs.

    \item {\bf Experimental result reproducibility}
    \item[] Question: Does the paper fully disclose all the information needed to reproduce the main experimental results of the paper to the extent that it affects the main claims and/or conclusions of the paper (regardless of whether the code and data are provided or not)?
    \item[] Answer: \answerYes{}
    \item[] Justification: Sec.~4.1 specifies all key hyperparameters ($K{=}4$, $N{=}64$, resolution $512{\times}384$, window radius $\Delta$, SAM3 threshold 0.5, agreement threshold $\tau{=}0.7$), the backbone models (Qwen3-VL-8B/30B), and the hardware platform (8$\times$ H100); Appendix Sec.~3 provides further experimental details.

\item {\bf Open access to data and code}
    \item[] Question: Does the paper provide open access to the data and code, with sufficient instructions to faithfully reproduce the main experimental results, as described in supplemental material?
    \item[] Answer: \answerNo{}
    \item[] Justification: Code will be released on the project page upon acceptance; the evaluation dataset SceneFun3D is publicly available at the benchmark website.

\item {\bf Experimental setting/details}
    \item[] Question: Does the paper specify all the training and test details (e.g., data splits, hyperparameters, how they were chosen, type of optimizer) necessary to understand the results?
    \item[] Answer: \answerYes{}
    \item[] Justification: Sec.~4.1 covers dataset splits (split0: 30 val scenes, split1: 200 train scenes), evaluation metrics (AP$_{25}$, AP$_{50}$, AR$_{25}$, AR$_{50}$, mIoU), and all inference hyperparameters. UniFunc3D is fully training-free, so no optimizer or training details apply.

\item {\bf Experiment statistical significance}
    \item[] Question: Does the paper report error bars suitably and correctly defined or other appropriate information about the statistical significance of the experiments?
    \item[] Answer: \answerNo{}
    \item[] Justification: Error bars are not reported, which is standard practice for benchmark evaluation in 3D scene understanding; the evaluation covers the full test set (split0: 30 scenes, split1: 200 scenes) rather than random subsets, so single-run results are representative.

\item {\bf Experiments compute resources}
    \item[] Question: For each experiment, does the paper provide sufficient information on the computer resources (type of compute workers, memory, time of execution) needed to reproduce the experiments?
    \item[] Answer: \answerYes{}
    \item[] Justification: Appendix Sec.~3 specifies the hardware platform (Intel Xeon Platinum 8468, 8$\times$ NVIDIA H100 GPUs), and Appendix Sec.~1 reports per-scene inference time ($\sim$26 min for UniFunc3D vs.\ $\sim$82 min for Fun3DU) on this platform.
    
\item {\bf Code of ethics}
    \item[] Question: Does the research conducted in the paper conform, in every respect, with the NeurIPS Code of Ethics \url{https://neurips.cc/public/EthicsGuidelines}?
    \item[] Answer: \answerYes{}
    \item[] Justification: The research involves 3D scene understanding on a publicly released benchmark dataset and does not involve human subjects, sensitive data collection, or harmful applications.

\item {\bf Broader impacts}
    \item[] Question: Does the paper discuss both potential positive societal impacts and negative societal impacts of the work performed?
    \item[] Answer: \answerNo{}
    \item[] Justification: The paper does not include a broader impact statement. The research advances embodied AI for indoor scene understanding, with clear positive applications in assistive robotics; we see no direct path to significant negative applications beyond general misuse risks common to perception systems.
    
\item {\bf Safeguards}
    \item[] Question: Does the paper describe safeguards that have been put in place for responsible release of data or models that have a high risk for misuse (e.g., pre-trained language models, image generators, or scraped datasets)?
    \item[] Answer: \answerNA{}
    \item[] Justification: UniFunc3D is a training-free inference pipeline that uses publicly available frozen models (Qwen3-VL, SAM3); no new model weights are trained or released, and the evaluation dataset is an existing public benchmark.

\item {\bf Licenses for existing assets}
    \item[] Question: Are the creators or original owners of assets (e.g., code, data, models), used in the paper, properly credited and are the license and terms of use explicitly mentioned and properly respected?
    \item[] Answer: \answerYes{}
    \item[] Justification: All datasets (SceneFun3D), models (Qwen3-VL, SAM3, Qwen2.5-VL), and baselines (Fun3DU, AffordBot, TASA, OpenMask3D, LERF, OpenIns3D) are properly cited with their original papers; all models used are publicly released research models.

\item {\bf New assets}
    \item[] Question: Are new assets introduced in the paper well documented and is the documentation provided alongside the assets?
    \item[] Answer: \answerNo{}
    \item[] Justification: The paper introduces code for the UniFunc3D inference pipeline, which will be released upon acceptance; documentation is not yet available at submission time.

\item {\bf Crowdsourcing and research with human subjects}
    \item[] Question: For crowdsourcing experiments and research with human subjects, does the paper include the full text of instructions given to participants and screenshots, if applicable, as well as details about compensation (if any)?
    \item[] Answer: \answerNA{}
    \item[] Justification: The paper does not involve crowdsourcing or research with human subjects; evaluation is conducted entirely on an existing annotated benchmark dataset.

\item {\bf Institutional review board (IRB) approvals or equivalent for research with human subjects}
    \item[] Question: Does the paper describe potential risks incurred by study participants, whether such risks were disclosed to the subjects, and whether Institutional Review Board (IRB) approvals (or an equivalent approval/review based on the requirements of your country or institution) were obtained?
    \item[] Answer: \answerNA{}
    \item[] Justification: The paper does not involve human subjects; no IRB approval is required.

\item {\bf Declaration of LLM usage}
    \item[] Question: Does the paper describe the usage of LLMs if it is an important, original, or non-standard component of the core methods in this research? Note that if the LLM is used only for writing, editing, or formatting purposes and does \emph{not} impact the core methodology, scientific rigor, or originality of the research, declaration is not required.
    \item[] Answer: \answerYes{}
    \item[] Justification: Multimodal LLMs (Qwen3-VL-8B and 30B, and Qwen2.5-VL-72B in ablations) are the central component of UniFunc3D's method; their role in active spatial-temporal grounding, functional object identification, and mask verification is described in detail throughout Sec.~3.

\end{enumerate}

\end{document}